\documentclass[runningheads]{llncs}

\usepackage{eccv}

\usepackage{eccvabbrv}
\usepackage{multirow}
\usepackage{graphicx}
\usepackage{booktabs}

\usepackage[accsupp]{axessibility}  

\usepackage{hyperref}

\usepackage{orcidlink}

\begin{document}

\title{Assessing Sample Quality via the Latent \\ Space of Generative Models} 


\author{Jingyi Xu\inst{1} \and
Hieu Le\inst{2} \and
Dimitris Samaras\inst{1}}

\authorrunning{J.~Xu et al.}

\institute{Stony Brook University, New York, USA \and EPFL, Lausanne, Switzerland}

\maketitle

\begin{abstract}
Advances in generative models increase the need for sample quality assessment. To do so, previous methods rely on a pre-trained feature extractor to embed the generated samples and real samples into a common space for comparison. However, different feature extractors might lead to inconsistent assessment outcomes.  Moreover, these methods are not applicable for domains where a robust, universal feature extractor does not yet exist, such as medical images or 3D assets. In this paper, we propose to directly examine the latent space of the trained generative model to infer generated sample quality. This is feasible because the quality a generated sample directly relates to the amount of training data resembling it, and we can infer this information by examining the density of the latent space. Accordingly, we use a latent density score function to quantify sample quality. We show that the proposed score correlates highly with the sample quality for various generative models including VAEs, GANs and Latent Diffusion Models. Compared with previous quality assessment methods, our method has the following advantages: 1) pre-generation quality estimation with reduced computational cost, 2) generalizability to various domains and modalities, and 3) applicability to latent-based image editing and generation methods. 
Extensive experiments demonstrate that our proposed methods can benefit downstream tasks such as few-shot image classification and latent face image editing. Code is available at \href{https://github.com/cvlab-stonybrook/LS-sample-quality}{https://github.com/cvlab-stonybrook/LS-sample-quality}.
\end{abstract}

\section{Introduction}
\label{sec:intro}


Generative models have emerged as powerful modeling tools that can capture diverse and complex distribution from a large training dataset to synthesize new data\cite{XuICCV21,Xu2022FS,Xu2023FSOD,Xu2023ZeroShotOC}. A single pre-trained diffusion model\cite{ldm} can generate thousands of images of ``{Yorkshire Terrier}'' or ``{Notre-Dame de Paris}''. In this paper, we aim to answer the question: among the samples generated from the model, how to measure the quality of each individual one? Such an instance-wise quality assessment metric is essential for users and consumers to select samples among the ones provided by those recently released text-to-image models, \textit{e.g.}, DALL-E 2 \cite{dalle-2} and Stable Diffusion \cite{ldm}, rather than model-wise metrics such as Frech\`et Inception Distance (FID) \cite{fid}. 

For the most part, previous instance-wise evaluation methods \cite{realismscore, Han2023rarity} rely on a pre-trained feature extractor (\textit{e.g.}, VGG16 \cite{VGG16}) to embed the generated samples and real samples into a common feature space. $K$-nearest neighbor ($k$-NN) based approaches are then applied under the assumption that close samples in this feature space correspond to semantically similar images.
The realism score \cite{realismscore}, for example, measures the maximum of the inverse relative distance of a fake sample in a real $k$-NN latent sphere. The rarity score \cite{Han2023rarity}, on the other hand, measures the minimum radius of a real $k$-NN sphere that contains the fake latent representation. However, relying on a pre-trained feature extractor suffers from two shortcomings. First, different feature extractors might lead to inconsistent assessment outcomes: the rarity score shows a negative correlation with Frech\`et Inception Distance (FID) \cite{fid} when using VGG16 as backbone, while the correlation becomes positive under DINO \cite{Caron2021Dino} or CLIP \cite{Radford2021LearningTV_CLIP} backbones. Moreover, these methods are not applicable for domains where a robust, universal feature extractor is not yet available, \textit{e.g.}, 3D shapes, human-drawn art or medical images.

In this paper, we propose to assess sample quality from another perspective: instead of using a pre-trained feature space, we directly use\textit{ the latent space of the generative models themselves}. 
The intuition is that the quality of a generated sample directly relates to the amount of the training samples that closely resemble it, and we can infer this information solely by examining the density of the latent space. 
Specifically, the samples lying in the latent area with dense latent codes
are likely to have sufficient training data resembling them while low-density latent areas would correspond to the rare cases in the data manifold. 
This is because generative models typically map similar data points to similar latent embeddings. 
The latent embeddings in those low-density areas are less exposed in model training, consequently receiving less supervision, and leading to potentially inferior reconstruction quality.

To this end, we propose a latent density score function to measure the quality of generated samples. Given a pre-trained generative model, our proposed function quantitatively measures the density of a randomly sampled latent code w.r.t. a set of latent codes extracted from the training data. 
We show that the proposed latent density score highly correlates with the sample quality for various generative models including Variational Autoencoders (VAEs) \cite{vae}, Generative Adversarial Networks (GANs) \cite{gan} and Latent Diffusion Models (LDMs) \cite{ldm}.
Compared with previous quality assessments that require an additional embedding network for feature extraction, our method estimates the sample quality by directly examining the latent space of the generative models, which brings several key advantages: 
\iftrue
1) \textbf{efficiency}: our method enables quality assessment without generating image pixels, which significantly reduces the computational cost;
2) \textbf{generalizability}: our method eliminates the reliance on external feature extractors, which allows for generalization to the domains where a universal pre-trained feature extractor might not exist;
3) \textbf{applicability}: our method can be seamlessly incorporated into latent-based image editing and generation methods, which can benefit various downstream tasks. 

In short, our main contributions can be summarized as follows:
\begin{itemize}
\setlength\itemsep{0.5em}
    \item We demonstrate that we can directly assess sample quality via the latent space of generative models themselves, while previous quality assessment methods rely on a pre-trained feature extractor to embed real and generated samples to a common space. 
    \item We propose a score function to quantify sample quality by measuring the density in latent space. The proposed function is applicable to various generative models trained on a variety of datasets. 
    \item We show the clear advantages of our proposed method over previous instance-wise evaluation methods, including significantly saving computational cost, generalizing across different domains and facilitating various downstream tasks.
\end{itemize}

\section{Related Work}
Previous metrics for quality assessment can be grouped into two main categories: model-wise evaluation metrics and instance-wise evaluation metrics. Model-wise evaluation metrics measure the performance of different generative models, while instance-wise evaluation metrics aim to compare the quality of each individual generated sample.

\subsection{Model-wise Evaluation Metrics} 
Various model-wise evaluation metrics have been proposed to quantify the performances of generative models.
Prevalent model-wise metrics include Inception Score (IS) \cite{is}, Kernel Inception Distance (KID) \cite{kid} and Frech\`et Inception Distance (FID) \cite{fid}.
They quantify the performance of a generative model by measuring the distribution discrepancy between the generated samples and real samples in a high-dimensional feature space. 
Sajjadi \textit{et al.} \cite{Sajjadi2018AssessModel} propose to further disentangle this discrepancy between distributions into two components: precision and recall. Precision represents the quality of generated samples while recall corresponds to the coverage of the real target distribution. 
Naeem \textit{et al.}  \cite{Naeem2020density} improve upon precision and recall by introducing density and coverage: density improves upon precision by being more robust to outliers and coverage improves upon recall by preventing the overestimation of the latent manifold.
Although the above metrics have demonstrated their effectiveness in assessing generative models, they are not suitable to measure individual sample quality since they work on a set of generations.

\subsection{Instance-wise Evaluation Metrics} 
Unlike model-wise metrics, instance-wise metrics are applied on individual generated samples for performance evaluation.
They are helpful for users to select samples from generative models, which might produce noisy, unrealistic samples with artifacts, especially for underrepresented cases \cite{self_diagnosing_gan} such as rare categories or extreme object poses. 
The realism score \cite{realismscore} measures the perceptual quality of individual samples by estimating how close a given fake sample is to the latent manifold of real samples.
Recently, Han \textit{et al.} have proposed the rarity score \cite{Han2023rarity}, which measures how rare a synthesized sample is based on the real data distribution.
Our proposed method and rarity score share the spirit of estimating the density around the target fake sample on the real manifold.
Nevertheless, rarity score defines this manifold using a pre-trained classification network, \textit{e.g.}, VGG16, while our method directly leverages the latent manifold of the generative models themselves. 
We show that in this latent manifold, the density 
correlates with the perceptual quality of the generated samples.

\section{Latent Density Score}
\label{sec:method}
Given a well-trained generative model, \textit{e.g.}, GAN, VAE or latent diffusion model, we aim to estimate the quality of the generated samples by examining the latent space of the model. Let $ \mathcal{Z} = \{z_1, z_2, ...z_i\}$ denote a set of latent codes extracted from the training samples,  and $z_g$ denote a latent code randomly sampled from the latent space,  we measure the latent density of $z_g$
quantitatively by calculating the latent density score as:
\begin{equation}\label{eq:density}
    D ({z_g, \mathcal{Z}}) = \frac{1} {\lvert \mathcal{Z} \rvert} * \sum_{z_i \in \mathcal{Z}} e ^ {-\frac{{\lVert z_g - z_i \rVert}^2}{2\sigma^2}},
\end{equation}
where $\sigma$ is a hyper-parameter of this score function.
Latent density score measures the average Gaussian kernelized Euclidean distance \cite{mean-shift} between $z_g$ and each latent code in $ \mathcal{Z}$. The score is high when $z_g$ resides in an area where the trained codes are densely distributed.
$\sigma$ controls the relative contribution of each latent code in $ \mathcal{Z} $ to the final density value, \textit{i.e.}, using a small $\sigma$ places more emphasis on the local area surrounding $z_g$, while applying a large $\sigma$ places more focus on the global density. In the case where there are multiple local clusters in the latent manifold, different values of $\sigma$ will lead to different assessment results (see Section \ref{sec:sigma}). 

In GAN-based generative models, truncation trick \cite{stylegan,glow,largescale_gan} is a widely used technique to increase the sample fidelity at the cost of lowering the diversity. It works by shifting a randomly sampled code towards the mean latent code.
The mean code typically resides in a high-density latent area. 
In fact, we observe that the proposed latent density score well correlates with the degree of truncation. 
We analyze this correlation further in Section \ref{sec: relationship} and provide more qualitative results in the Supplementary Material.
Another highly relevant quality assessment metric is the realism score \cite{realismscore}. The realism score measures the relative distance of a fake sample in a real latent sphere, which is defined by a pre-trained feature extractor.
We show that the latent density score behaves similarly with the realism score for images from the domains previously seen by the feature extractor (see Section \ref{sec: relationship}). However, for images from non-ImageNet-like domains (\textit{e.g.}, medical images and anime-style images) or domains other than 2D images (\textit{e.g.}, 3D shapes), quality assessment with realism score will be infeasible (see Section \ref{sec:cross_domain}).

\begin{figure*}[]
\begin{center}
\subfloat[Latent Diffusion - CelebA-HQ]{%
  \includegraphics[clip,width=1\columnwidth]{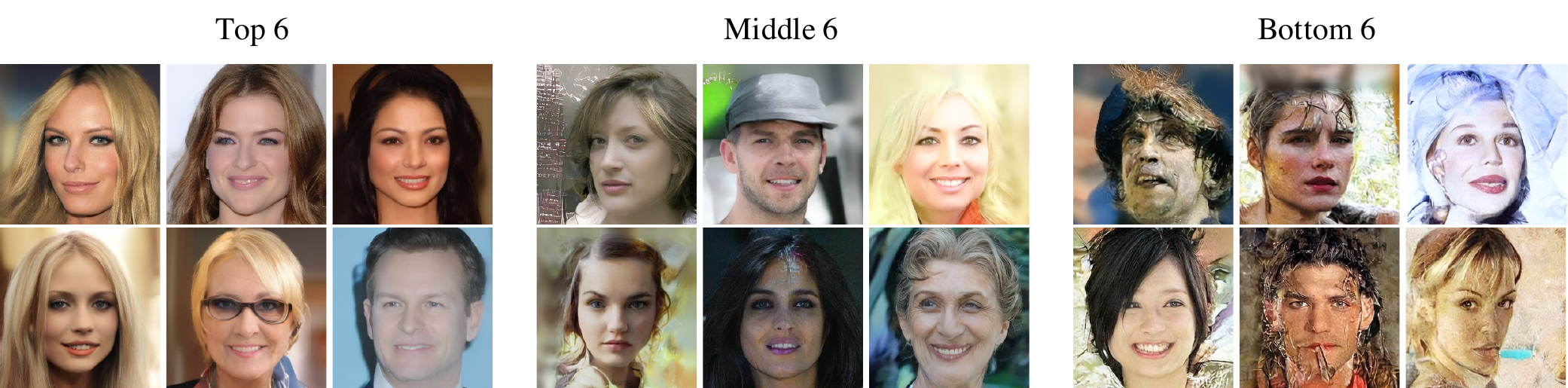}
  }
  \vspace{2mm}
\subfloat[Latent Diffusion - LSUN-Bedrooms]{%
  \includegraphics[clip,width=1\columnwidth]{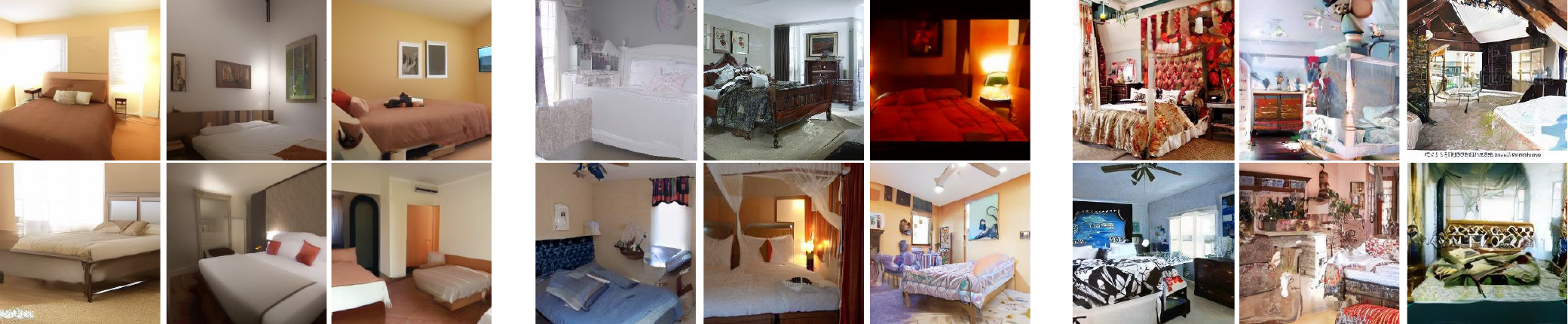}%
}
\vspace{2mm}
\subfloat[Latent Diffusion - LSUN-Churches]{%
  \includegraphics[clip,width=1\columnwidth]{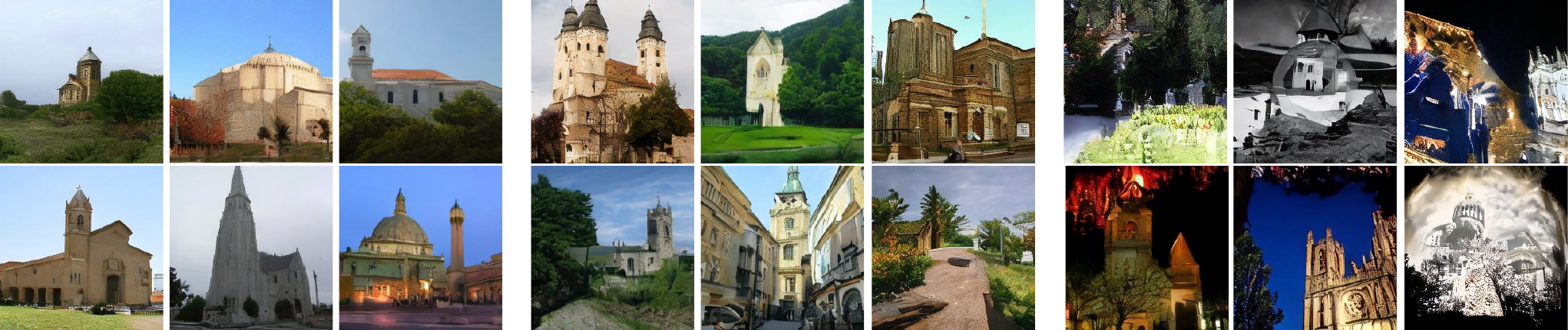}%
}
\caption{Top 6, Middle 6 and Bottom 6 generated images in terms of the proposed latent density score on CelebA-HQ, LSUN-Bedrooms and LSUN-Churches for unconditional latent diffusion models. (Zoom-in for best view). The proposed latent density scores highly correlate with the quality of generated images.} 
\vspace{0mm}
\label{fig: latent_diff_qual}
\end{center}
\end{figure*}

\begin{figure}
\begin{center}
\hspace{-0mm}\includegraphics[width=0.9\columnwidth]{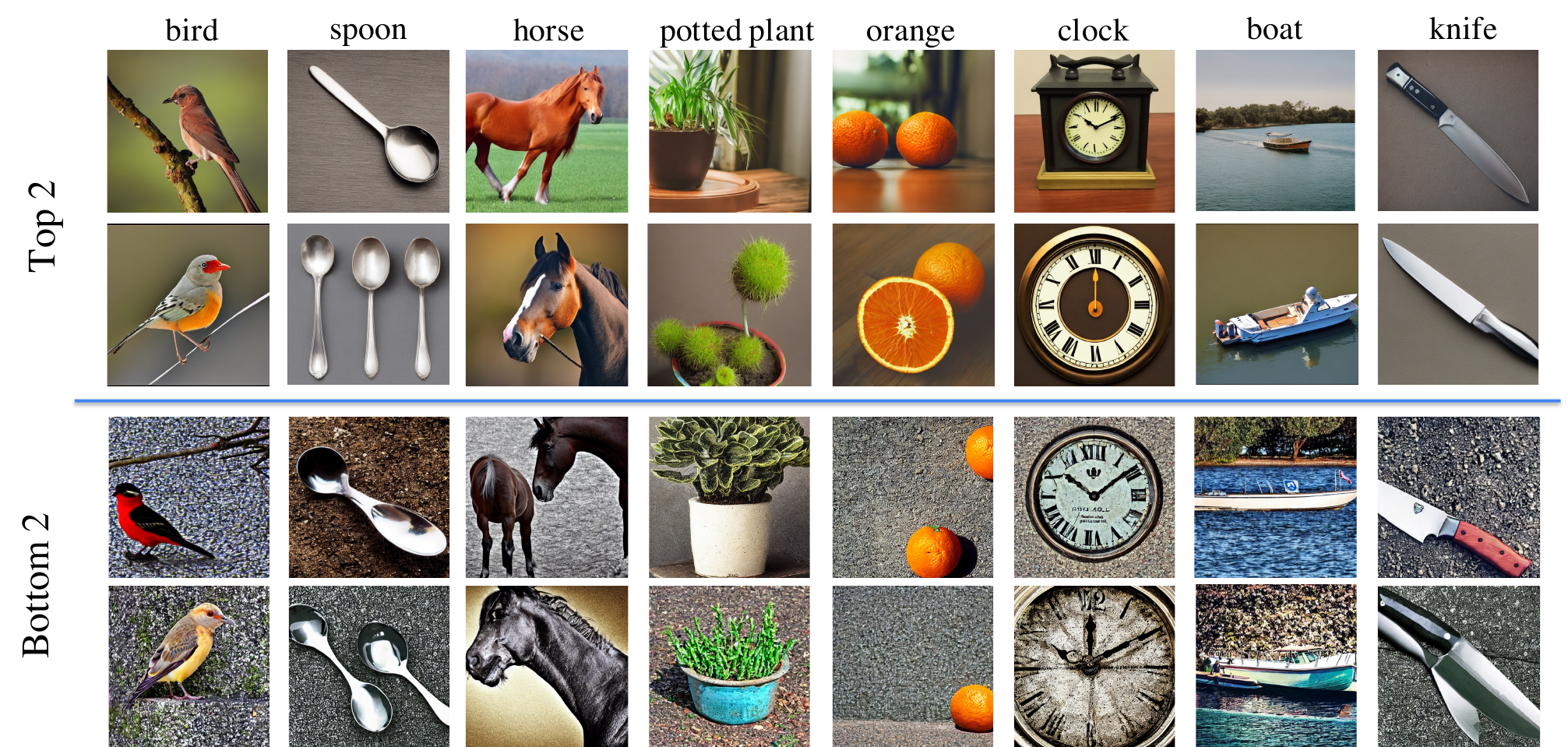} 
\caption{
Top 2 and bottom 2 Stable Diffusion generated samples for eight classes in terms of the proposed latent density score. Images in the `top 2' rows are high-resolution images with natural, realistic backgrounds, whereas images in the `bottom 2' rows contain visual noise and artifacts. 
The only difference in model configuration for images of top / bottom rows is the initial noise. 
} \vspace{-0mm}
\label{fig:sd_qual}
\end{center}
\end{figure}

\section{Experimental Results}
\label{sec:experiments}
\subsection{Results on Different Generative Models}
In this section, we provide the experimental results of the proposed metric for various generative models and datasets. We experiment with three types of generative models, \textit{i.e.}, GANs, VAEs and LDMs. 
For each trained model, we extract latent codes from 60k training samples and calculate the latent density scores for 20k randomly sampled latent codes. In particular, for VAEs and LDMs, we take the output of the image encoder as the latent representation of each real input image. 
For LDMs, we further flatten the 2D representations (before the denoising process) for computing the latent density score.
\begin{figure}
\begin{center}
\subfloat[VAE - MNIST]{%
\hspace{-8mm} \includegraphics[clip,width=0.8\columnwidth]{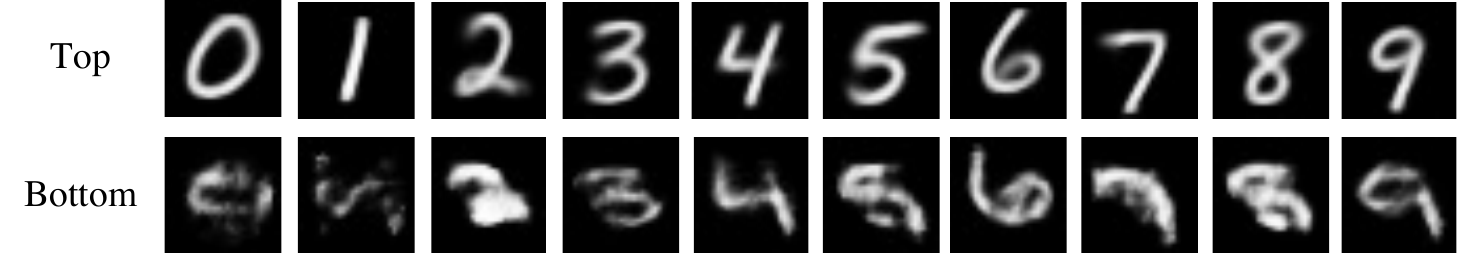}
  }\\
\subfloat[VAE - Fashion-MNIST]{%
\hspace{-8mm} \includegraphics[clip,width=0.8\columnwidth]{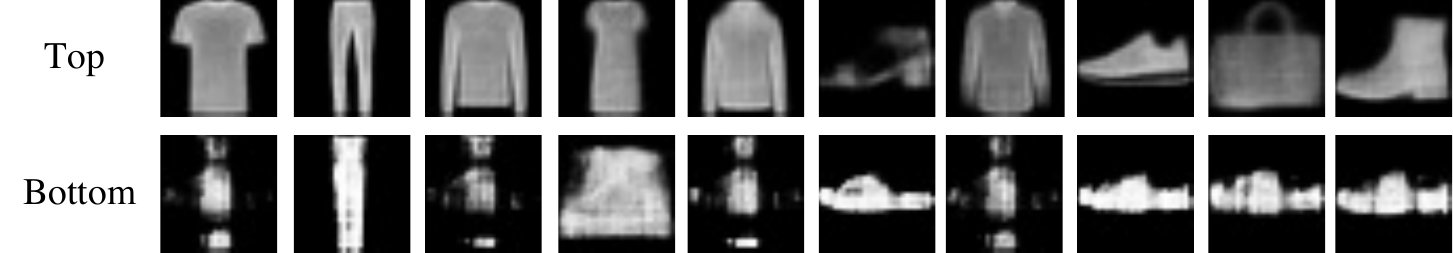}%
}\\
\subfloat[VAE - CelebA]{%
 \hspace{-8mm} \includegraphics[clip,width=0.8\columnwidth]{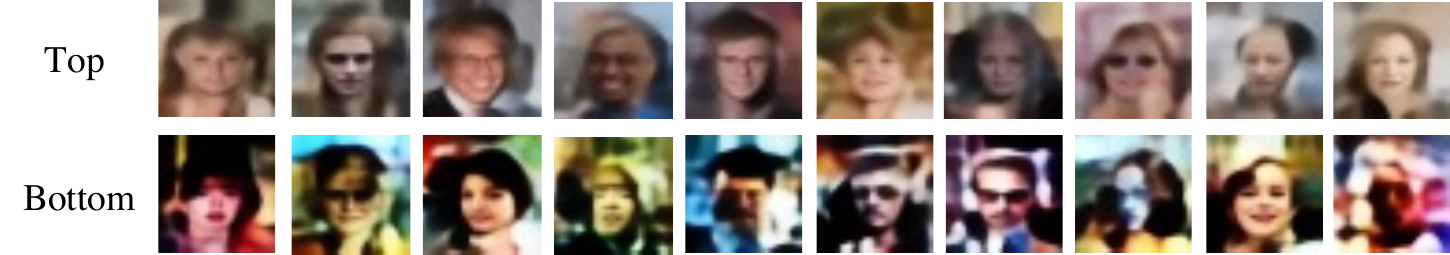}%
}
\caption{Top 10 and Bottom 10 generated images in terms of the proposed latent density score on MNIST, Fashion-MNIST and CelebA for VAE. The samples with high latent density scores display clear instances, whereas those with low latent density scores are often distorted / blurred.}
\vspace{-0.cm}
\label{fig: mnist_vae}
\end{center}
\end{figure}
We use the pre-trained Stable Diffusion v1.5 model \cite{ldm} as the text-to-image diffusion model. For the unconditional diffusion models, we choose the LDMs pre-trained on CelebA-HQ \cite{stargan}, LSUN-Bedrooms and LSUN-Churches \cite{lsun} released by \cite{ldm}.
For GANs, we experiment with StyleGAN2 \cite{stylegan2} and StyleGAN2-ADA \cite{stylegan2-ada}.
To obtain the latent representations, we input vectors sampled from a normal distribution to their mapping networks and extract latent features from the $\mathcal{W}$-space.
We use $\sigma=20$ for computing latent density scores. We analyze  the choice of $\sigma$ and how it affects the results in Section \ref{sec:sigma}.

\subsubsection{Latent Diffusion Models}
Latent diffusion models \cite{ldm} use pre-trained autoencoders to construct a low-dimensional latent space, from which the original data can be reconstructed at high fidelity with reduced computational costs. 
In Figure \ref{fig: latent_diff_qual}, we show images synthesized by unconditional latent diffusion models trained on CelebA-HQ, LSUN-Bedrooms and LSUN-Churches. For each dataset, we show samples using latent codes with the top 6 highest, top 6 lowest and 6 middle latent density scores. As shown in the figure, the proposed latent density scores highly correlate with the quality of generated images. For example, on the CelebA-HQ dataset, we can see human faces generated from codes with high latent density scores are visually realistic with clear hair, eye and eyebrow details, whereas those with low latent density scores are of degraded quality due to blur, artifacts or distorted facial structures.
Similarly, on LSUN-Bedrooms and LSUN-Churches, we observe unrealistic artifacts (\textit{i.e.}, distorted textures or inharmonious colors) from images with low latent density scores.

Figure \ref{fig:sd_qual} shows images synthesized by a pre-trained text-to-image diffusion model, \textit{i.e.}, Stable Diffusion, using latent codes with the top 2 highest and lowest latent density scores from eight classes. As shown in the figure, samples with high latent density scores have superior visual quality while latent codes with low scores often lead to erroneous samples. The most obvious failures are the unrealistic backgrounds. For example, the boat images in the `top 2' rows are high-resolution images with natural, realistic backgrounds, whereas the backgrounds of the boats in the `bottom 2' rows contain visual noise and artifacts. 
In some other failure cases, the generated objects exhibit structural integrity artifacts, \textit{i.e.}, the spoons and the clocks. 
We note that all images here are generated with the same model configuration and the only difference is the initial noise.
Previous works \cite{Dhariwal2021diffusion,Ho2021guidance,nichol2021glide} have shown that the guidance methods used in the denoising process are essential for improving the quality of generated images.
Here we observe that the randomly initialized noise can also affect the sample quality significantly.

\subsubsection{VAEs and GANs}

Figure \ref{fig: mnist_vae} presents images generated by VAEs using codes with the top 10 highest and top 10 lowest latent density scores. As shown in the figure, our proposed score function is applicable to VAEs as well.
For MNIST \cite{mnist} and Fashion-MNIST \cite{xiao2017fashionmnist}, for example, we observe the samples with high latent density scores display clear instances from the given class, whereas those with low latent density scores are often distorted / blurred to unrecognizable.

Figure \ref{fig: stylegan_qual} shows images generated by StyleGAN2 \cite{stylegan2} trained on FFHQ \cite{stylegan}, StyleGAN2-ADA \cite{stylegan2-ada} trained on AFHQ Dog \cite{stargan} and StyleGAN2 trained on AFHQ Cat \cite{stargan} using codes with the top 6 highest, top 6 lowest and 6 middle latent density scores. We observe clear generation quality differences between samples with different scores. For example, on the AFHQ Dog dataset, the samples with high scores show clear, frontal dog faces. On the other hand, the dog faces in samples with low density scores are highly distorted in various ways.  

\begin{figure*}
\begin{center}
\subfloat[StyleGAN2 - FFHQ]{%
  \includegraphics[clip,width=1\columnwidth]{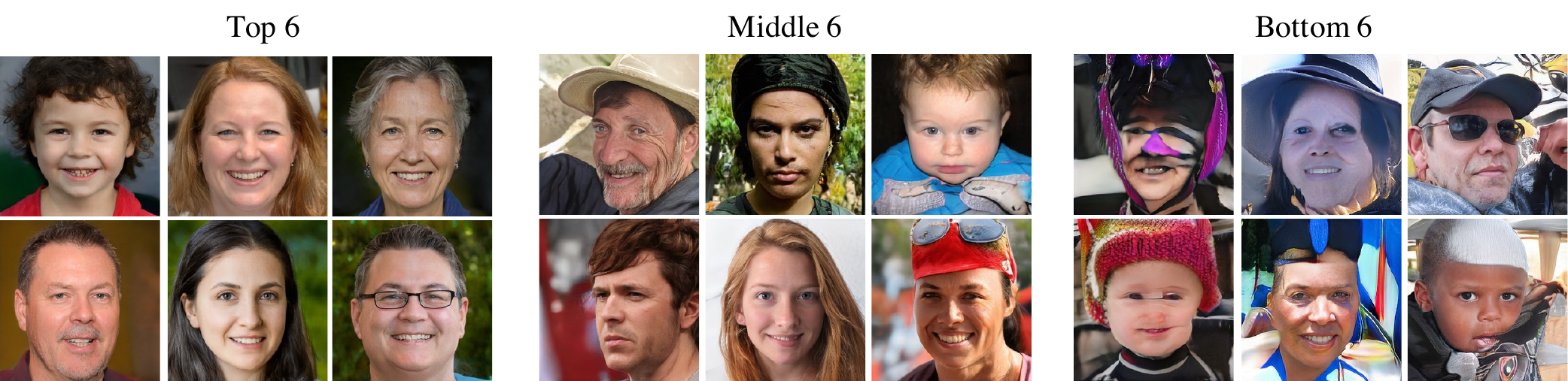}%
}
\vspace{3mm}
\subfloat[StyleGAN2-ADA - AFHQ Dog]{%
  \includegraphics[clip,width=1\columnwidth]{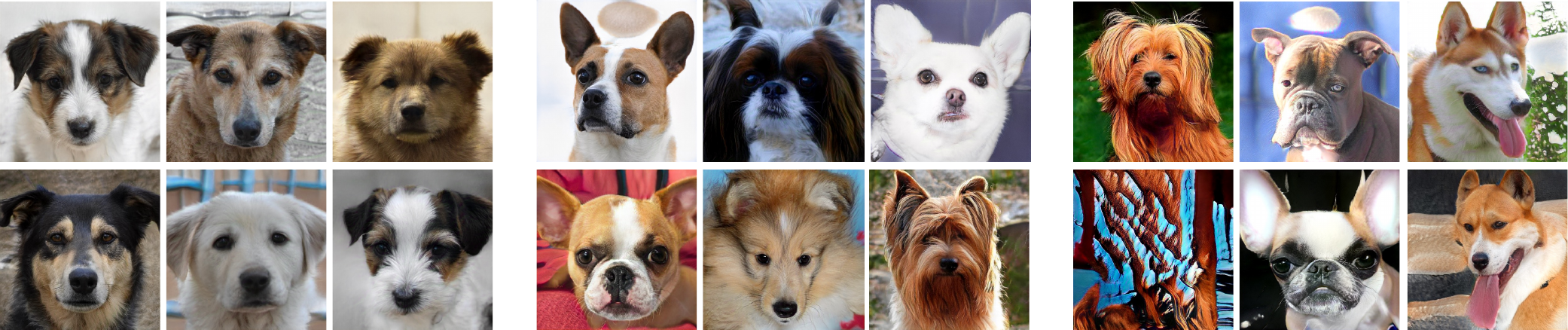}%
}
\vspace{3mm}
\subfloat[StyleGAN2 - AFHQ Cat]{%
  \includegraphics[clip,width=1\columnwidth]{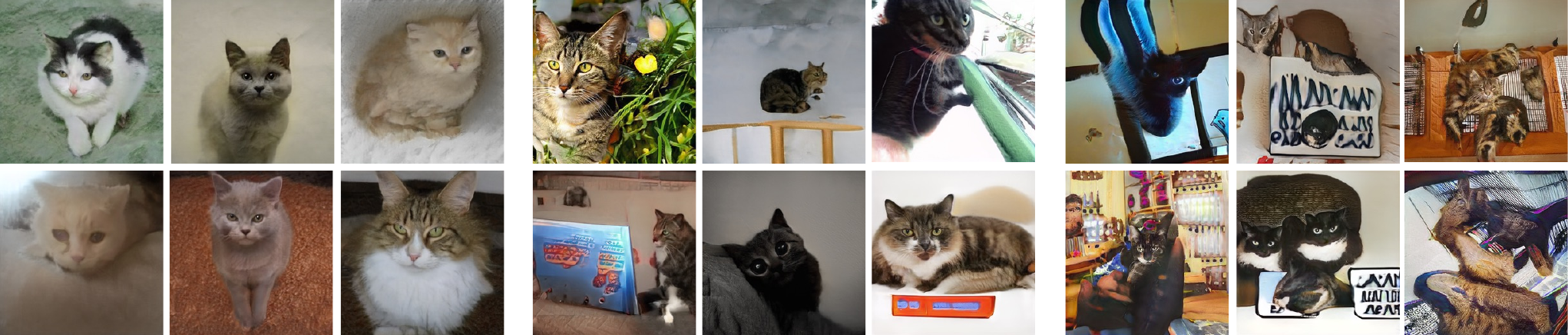}
 }
\caption{Top 6, Middle 6 and Bottom 6 generated images in terms of the proposed latent density score on FFHQ for StyleGAN2, on AFHQ Dog for StyleGAN2-ADA and on AFHQ Cat for StyleGAN2.  (Zoom-in for best view). Samples with high scores are of better quality while samples with low scores are often highly distorted.}
\label{fig: stylegan_qual}
\end{center}
\end{figure*}

\subsection{Results on Other Domains and Modalities}
\label{sec:cross_domain}


Our proposed metric does not rely on any additional feature extractor, which enables quality assessment in the domains where robust pre-trained models might not be available. In this section, we show the applications of our method on quality assessment for generated 3D shapes and non-ImageNet-like images.

\subsubsection{Quality Assessment for 3D Shapes}
\begin{figure*}
\begin{center}
\vspace{-0.2mm}
{%
  \includegraphics[clip,width=\columnwidth]{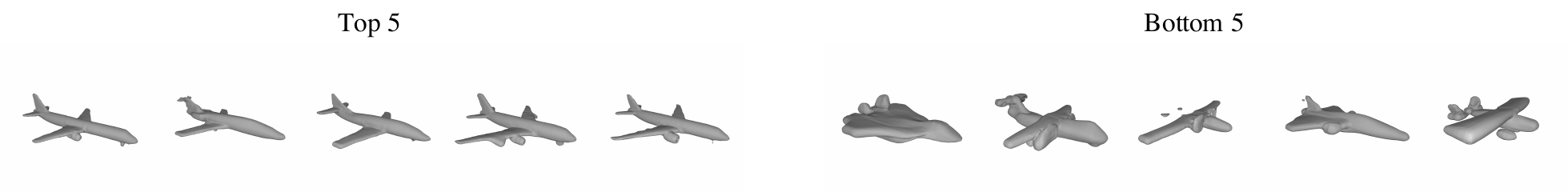}%
}
\vspace{0.3mm}
{%
  \includegraphics[clip,width=\columnwidth]{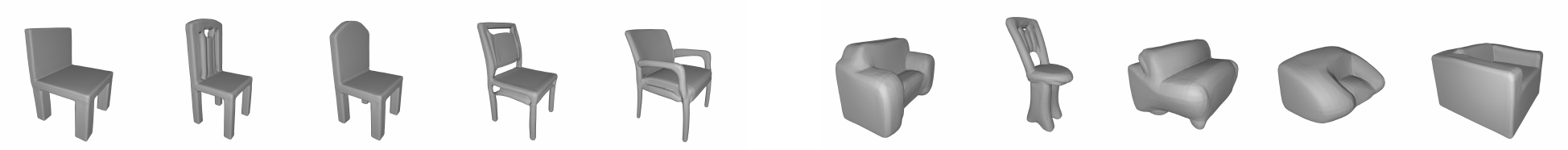}%
}
\vspace{0.3mm}
{%
  \includegraphics[clip,width=\columnwidth]{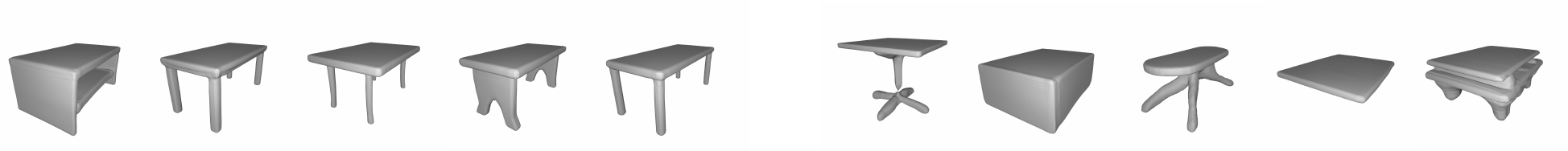}%
}
\vspace{0.3mm}
{%
  \includegraphics[clip,width=\columnwidth]{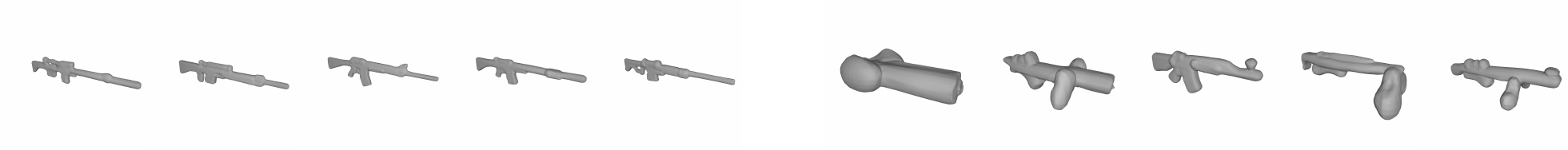}%
}
\caption{Top 5 and Bottom 5 generated 3D shapes for four categories (\textit{i.e.}, airplane, chair, table and rifle) in terms of the proposed latent density score on ShapeNet Core V1 for SDF-StyleGAN. The generated samples with high scores have plausible 3D shapes and complete geometry structures, while samples with low scores exhibit unrealistic shapes and severe geometry distortion.}
\label{fig: 3d}
\vspace{-0mm}
\end{center}
\end{figure*}
We first show the application of our method on generated 3D shapes. Specifically, we generate shapes for four categories, \textit{i.e.}, airplane, chair, table and rifle, using a StyleGAN2-based 3D shape generation framework, SDF-StyleGAN \cite{zheng2022sdfstylegan} trained on ShapeNet Core V1 \cite{shapenet}.
For each shape category, we extract the latent embeddings in the $\mathcal{W}$ space of SDF-StyleGAN for 30k randomly sampled vectors and compute the corresponding latent density scores. Figure \ref{fig: 3d} visualizes the generated shapes with the top 5 highest and lowest scores.
We observe that the generated 3D shapes with high scores have better visual quality with plausible 3D shapes and complete geometry structures. The generated 3D shapes with low scores, in contrast, exhibit unrealistic shapes and severe geometry distortion.

\subsubsection{Quality Assessment on Non-ImageNet-like Images}
\label{sec:cross_domain_images}
\begin{figure}
\begin{center}
\subfloat[StyleGAN2 - Danbooru]{%
\hspace{-4.5mm} \includegraphics[clip,width=0.95\columnwidth]{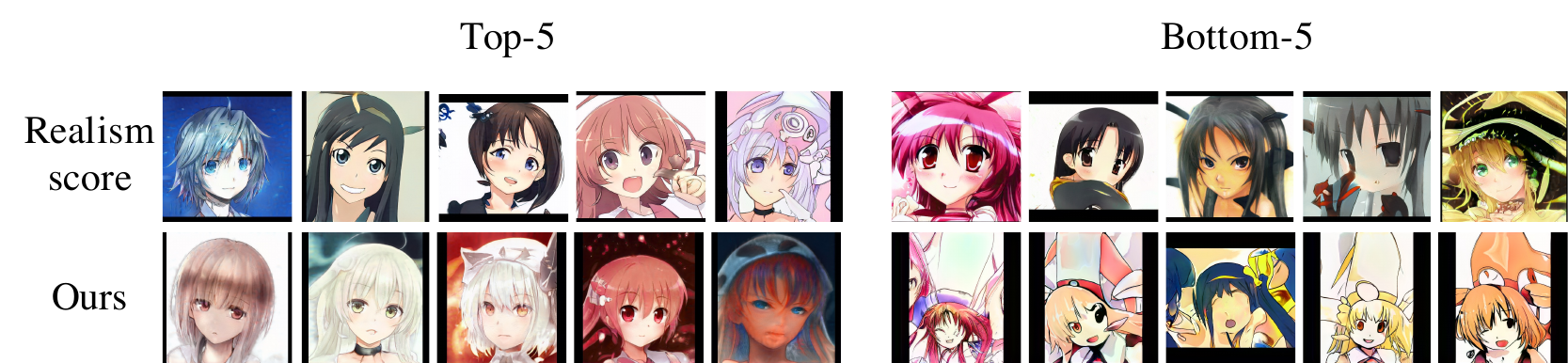}
  }
\vspace{4mm}
\subfloat[StyleGAN2 - BreCaHAD]{%
 \hspace{-4.5mm} 
 \includegraphics[clip,width=0.95\columnwidth]{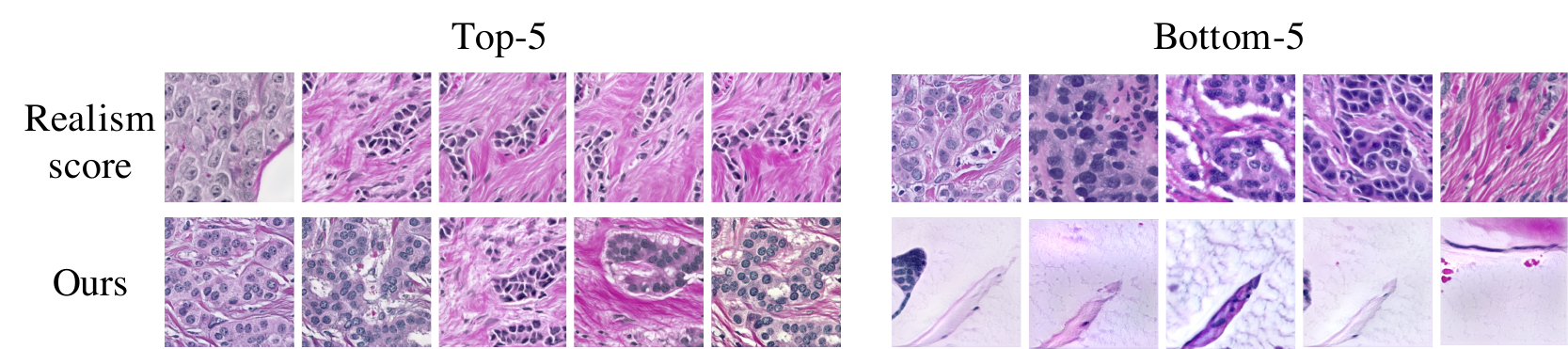}%
}
\caption{Top and bottom generated images in terms of latent density score and realism score using StyleGAN2 pre-trained on BreCaHAD \cite{Aksac2019brecahad} and Danbooru \cite{danbooru} datasets. There is clear visual difference between images with the highest and lowest latent density scores. In comparison, we do not observe visually distinguishable difference between samples with the highest and lowest realism scores. }
\vspace{-2.3em}
\label{fig: domain}
\end{center}
\end{figure}

Existing quality assessment methods operate under the assumption that semantically similar images are mapped to points close to each other in the embedding space of a pre-trained feature extractor.
However, this assumption might not hold true across different data domains.
In this section, we conduct quality assessment for images from two non-ImageNet-like domains, \textit{i.e.}, the  medical domain and anime-style domain. Figure \ref{fig: domain} shows the samples with the highest and lowest latent density scores / realism scores among 5k candidate samples on each domain. We can see that the images with the highest and lowest latent density scores exhibit clear visual difference. On the BreCaHAD dataset \cite{Aksac2019brecahad}, for example, the high-density images contain representative human cells while low-density images are mostly blank. 
We note that the low-density samples do not show degraded perceptual quality. This is probably because although these images are underrepresented cases in the training set, reconstructing them is relatively easy due to their simple layouts.

On the other hand, we do not observe visually distinguishable differences between samples with the highest and lowest realism scores. This suggests that the pre-trained VGG space used by the realism score is not semantically meaningful for non-ImageNet-like domain images.
In addition, our method is more computationally efficient, since we directly operate on the latent codes instead of actually generating all the 5k candidate images.

\section{Applications}
\label{sec:applications}
\subsection{Latent Face Editing}

 \begin{figure}[h!]
 \centering
\includegraphics[width=1\linewidth]{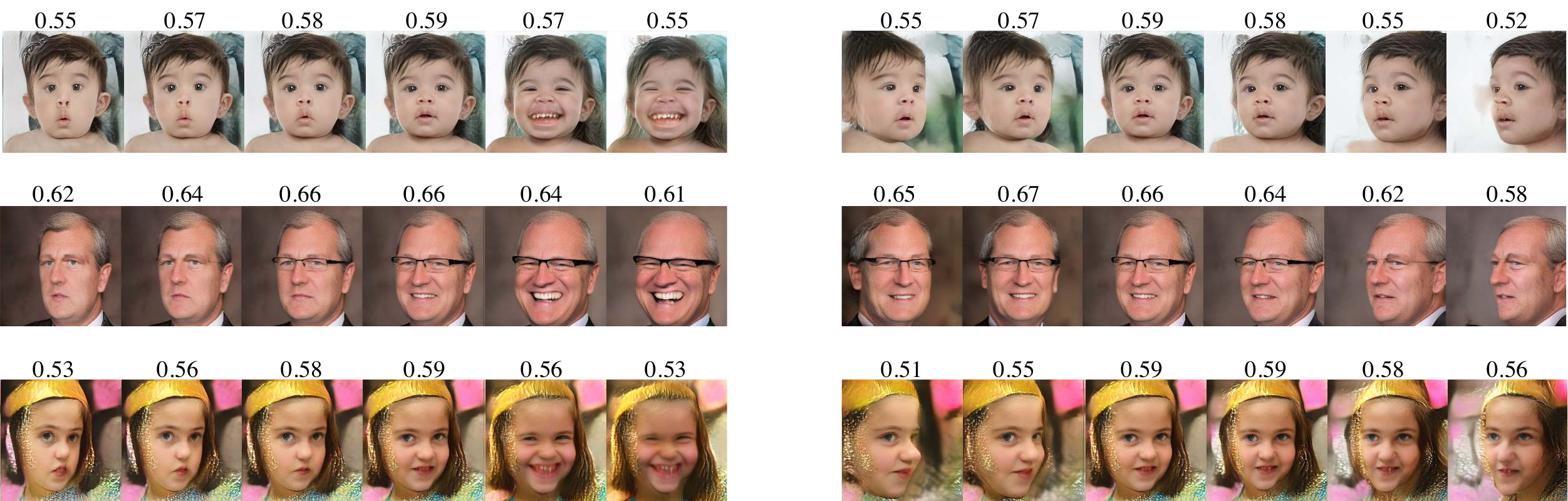}
 \caption{ {Latent-based face editing results and the corresponding latent density scores.} The latent density scores highly correlate with the quality of the images.}
\label{fig:face_edit}
\vspace{0.cm}
\end{figure}

Our proposed method operates directly on the latent space of the generator. Thus, it can be seamlessly incorporated into latent-based image editing methods. Previous work \cite{interfacegan} has shown that by moving a latent code along certain directions in the latent space of a well-trained face synthesis model, one could control facial attributes of the generated images. However, if the code is moved too far from the well-behaved regions \cite{goodlatents} of the latent space, the generated samples will suffer from severe changes \cite{interfacegan} as well as degradation in image quality. Here we use our method to estimate the perceptual quality of the edited samples. Specifically, we take a latent code and move it along the direction for the attribute ``pose" in the latent space $\mathcal{W}$ of StyleGAN2 following \cite{interfacegan}. We compute the latent density score of the moved latent code based on Equation \ref{eq:density}.
Figure \ref{fig:face_edit} shows the generated edited images and the corresponding scores.

As shown in the figure, the latent density scores well correlate with the quality of the manipulated images: images with low scores contain artifacts while images with high scores are of better quality. Our method provides a reliable way to assess the quality of edited images even before generating them, which helps to avoid image corruption during latent space traversal and facilitates meaningful image manipulation.
\subsection{Few-Shot Image Classification} 
Our method enables selecting strictly high-quality generated images with clear, high-resolution objects. These images are particularly useful for augmenting the training set in low-shot scenarios\cite{Xu2022FS,Xu2023FSOD,Xu2023ZeroShotOC,XuICCV21}.
Here we show these samples can be used in the task of few-shot image classification and greatly boost performance. Specifically, we synthesize images using a pre-trained text-to-image model, \textit{e.g.}, Stable Diffusion, with the class name as the text condition. The synthesized images are then used as support samples for the corresponding class. We generate $k$ images for $k$-shot learning ($k = 1$ or $5$). For the feature extractor, we use ResNet12 \cite{tadam} trained following previous work \cite{metabaseline}.
Table \ref{tab:mini_tiered} compares the performance of using different sets of latent codes during image generation including: 1) $k$ randomly sampled codes, 2) top-$k$ codes with the highest, and 3) top-$k$ codes with the lowest latent density scores. Using samples with high latent density scores as support data leads to better few-shot performance on both the \textit{mini}ImageNet \cite{matchingnet} and \textit{tiered}ImageNet \cite{Ren2018MetaLearningFS_tieredImagenet} datasets for the $1$-shot and $5$-shot settings. In particular, results on $1$-shot \textit{mini}ImageNet show the largest margin, with a $3.28\%$ improvement over using random codes. This validates the superior quality of images generated from codes with high latent density scores.
\begin{table*}[t] 
    \caption{Few-shot image classification accuracy on \textit{mini}ImageNet and \textit{tiered}ImageNet using images generated from different sets of latent codes. Using images from the latent codes with highest latent density scores achieves better classification performance, which validates the superior quality of images with high latent density scores.}
    \label{tab:mini_tiered}%
    \centering
    \captionsetup{width=\textwidth}
  \resizebox{\textwidth}{!}{%
    \begin{tabular}{c|c|cc|cc}
      \hline
       & \multirow{2}{*}{Support Samples}  & \multicolumn{2}{c|}{\textit{mini}ImageNet} & \multicolumn{2}{c}{\textit{tiered}ImageNet}\\
      & & 1-shot & 5-shot & 1-shot & 5-shot  \\
      \midrule
       Real & -  &   63.17 $\pm$ 0.23 & 79.26 $\pm$ 0.17 & 68.62 $\pm$ 0.27 & 83.74 $\pm$ 0.18 \\
      \midrule
      \multirow{3}{*}{SD-generated} & bottom-$k$  & 63.43 $\pm$ 0.45 & 71.97 $\pm$ 0.37 & 64.62 $\pm$ 0.55 & 75.08 $\pm$  0.45 \\
      & random-$k$ &   63.87 $\pm$ 0.43 & 72.76 $\pm$ 0.38 & 66.04 $\pm$ 0.52 & 77.10 $\pm$ 0.44 \\
      & top-$k$ &  \textbf{67.15 $\pm$ 0.44} & \textbf{73.60 $\pm$ 0.37} & \textbf{68.39 $\pm$ 0.54} & \textbf{77.42 $\pm$ 0.43}    \\
      \hline
    \end{tabular}} 
    \vspace{0pt}
  \end{table*}

\section{Analysis}
\subsection{Relationship with Existing Metrics} 
\label{sec: relationship}

\begin{figure*}[h!]
\begin{center}
\includegraphics[clip,width=\columnwidth]{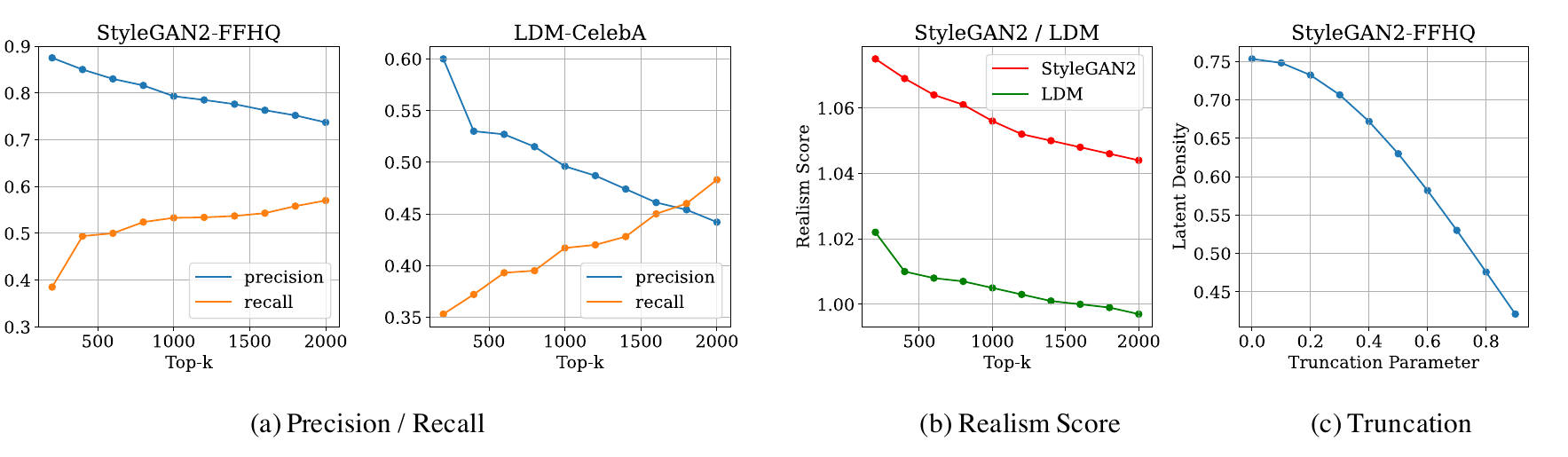}%
\vspace{-1mm}
\caption{Relationship between our proposed latent density score and other metrics. Top-$k$ samples are ranked according to our latent density score. The results of our proposed metric are aligned with existing evaluation metrics on images of common domains.}
\vspace{-0mm}
\label{fig: relation}
\end{center}
\end{figure*}

In this section, we investigate the relationship of our proposed metric with other existing metrics. In particular, we generate $2000$ fake samples and rank them based on the latent density score. Each time we select top-$k$ samples and calculate the corresponding precision / recall / realism scores of the selected samples. 
The scores of these metrics under different $k$ values are shown in Figure \ref{fig: relation} (a) and Figure \ref{fig: relation} (b). In addition, we show in Figure \ref{fig: relation} (c) how the latent density score changes when we increase the value of truncation parameter used in truncation trick. We conduct this experiment using StyleGAN2 trained on FFHQ \cite{stylegan} and LDM trained on CelebA-HQ \cite{stargan}.

\textbf{Precision and Recall}.
Precision and recall are commonly used evaluation metrics in many tasks, such as image classification or natural language processing.
In particular, precision measures the fraction of the generated samples that are realistic. Recall, on the other hand, measures the fraction of the real data distribution which can be covered by the distribution of fake data. 
As shown in Figure \ref{fig: relation} (a), a small value of $k$ leads to high precision and low recall. This suggests that the samples with high latent density scores are of high quality. 
As $k$ increases, more diverse samples are selected, which improves recall, while the decrease in precision indicates the newly selected samples are of inferior quality. 
The correlation between precision / recall and latent density score validates that our proposed metric reliably indicates sample quality.

\textbf{Realism Score. } The realism score is a highly relevant metric that measures the fidelity of an individual generated sample. As shown in Figure \ref{fig: relation} (b), as $k$ increases, the average realism score of the top-$k$ selected samples decreases for both StyleGAN2 and LDM. This suggests that our proposed metric is aligned with realism score, \textit{i.e.}, samples with low latent density scores also have low realism scores, and vice versa. However, the realism score relies on another feature extractor to project the generated samples to another space. Thus, it is not able to generalize to other domains or modalities (as shown in Section \ref{sec:cross_domain}). Moreover, computing realism score requires generating image data, which is time-consuming and not easily scalable, while our method directly operates on the latent space without the need for generating images. 
 
\textbf{Truncation Trick. } Truncation trick \cite{stylegan} is used to increase the fidelity of the generated images in GAN-based generative models by moving the latent code towards the mean latent code. The degree of truncation is controlled by the truncation parameter $\psi$, \textit{i.e.}, $\psi = 0$ indicates full truncation using the mean code and $\psi = 1$ indicates no truncation. We see from Figure \ref{fig: relation} (c) that the latent density score decreases as the truncation parameter increases. A higher degree of truncation typically leads to high-fidelity image generation, which, as we show, corresponds to a higher value of the latent density score. This suggests that the latent density score is a valid measure for generated sample quality.

\subsection{Effect of Hyper-Parameter $\sigma$}
\label{sec:sigma}
\begin{figure*}[h!]
  \centering
\begin{center}
\subfloat[]{%
\hspace{0mm}\vspace{5.5mm}\includegraphics[clip,width=0.7\columnwidth]{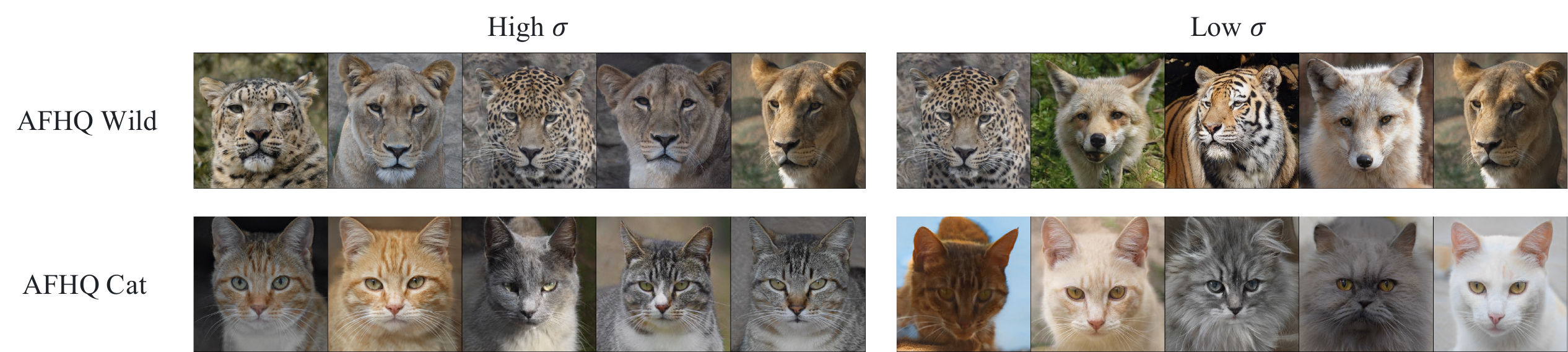}}
\subfloat[]{%
\includegraphics[clip,width=0.3\columnwidth]{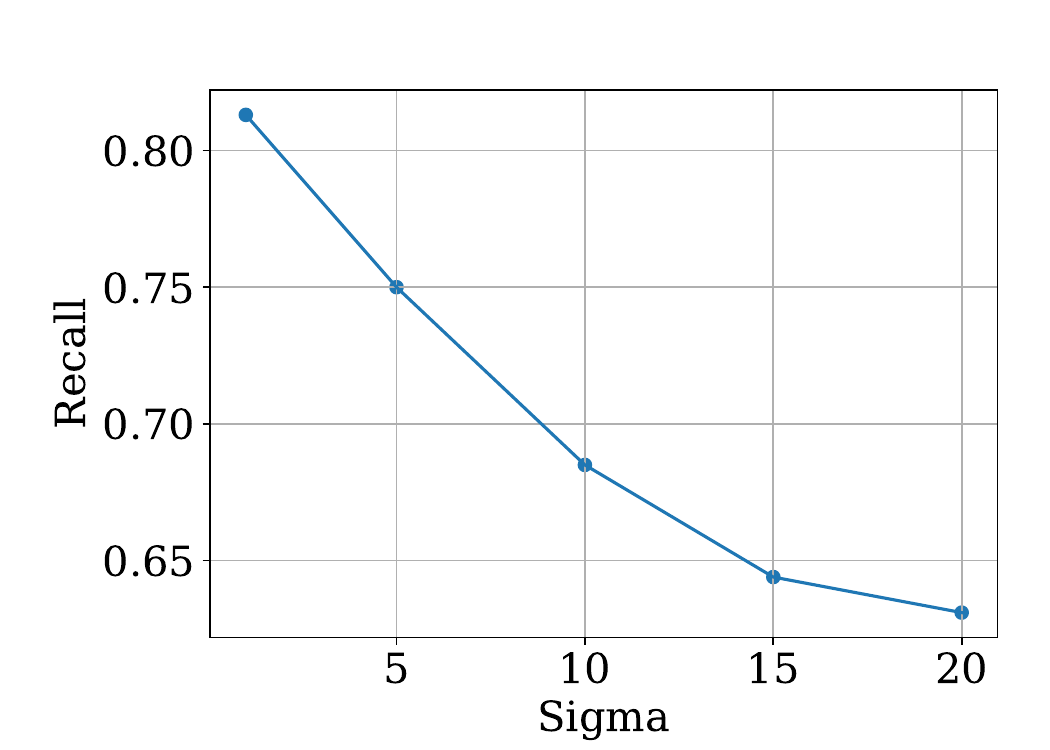}%
}
\caption{Images with high latent density scores using high and low $\sigma$ values on the AFHQ Wild and AFHQ Cat datasets (a) and different recall rates under different $\sigma$ values (b). We observe that under a lower $\sigma$ value, the images selected as high-density ones exhibit more diversity, which corresponds to a higher recall rate.}
\label{fig: sigma}
\end{center}
\end{figure*}

In this section, we analyze how the choice of $\sigma$ in Equation \ref{eq:density} affects quality assessment. Equation \ref{eq:density} measures the average Gaussian kernelized Euclidean distance between a given code and the latent codes extracted from training data, with $\sigma$ being the standard deviation of the kernel function. 
When applying a small $\sigma$, the final density value will rely relatively more on the area surrounding the given code.
This will increase the chance that points residing in local clusters are selected as high-density points. 
As a result, the selected samples are likely to be more diverse.
Figure \ref{fig: sigma} (a) shows the images with high latent density scores on the AFHQ Wild and AFHQ Cat datasets \cite{stargan} under large and small $\sigma$ values respectively. We observe that the selected high-density samples when using a small $\sigma$ are more diverse compared to using a large $\sigma$. Correspondingly, we observe a higher recall under a smaller $\sigma$ (as shown in Figure \ref{fig: sigma} (b)), indicating a more complete coverage of the latent manifold. $\sigma$ enables us to control the relative contribution of local density and global density w.r.t to the final density score. 
In this case, applying a small $\sigma$ allows us to select more diverse samples.

\section{Discussion and Conclusions}
In this paper, we have proposed a novel approach to estimate sample quality via the latent space of generative models. 
 Our method can be particularly useful in many scenarios. When training generative models, our proposed score points out the underrepresented cases that would possibly require collecting additional data\cite{li2024controllingfidelitydiversitydeep,Xu2022FS,XuICCV21,Xu2023FSOD,Xu2023ZeroShotOC,xu2023zeroshotobjectcountinglanguagevision,Le_2020_ECCV,le2020physicsbased,durasov2024enabling,durasov2024zigzag}. 
It also allows us to select high-quality samples that best benefit downstream tasks. 
For large-scale generative models, pre-generation quality assessment can greatly reduce computational costs.
However, only sampling data with high scores, might result in an incomplete coverage of the data manifold. This is because the scores are likely to be higher for large clusters of data representing common cases, as opposed to minority groups such as rare animal species\cite{Le_2019_CVPR_Workshops} or uncommon medical conditions. One way to alleviate this issue is by considering only small neighborhood areas when measuring the density, which can be achieved by applying a small value of $\sigma$. 
Further, previously proposed sampling techniques such as accept-reject sampling \cite{Dis2019azadi,li2024controllingfidelitydiversitydeep} can be used together with our method to increase sample diversity.
Combining our score with diversity-related scores also allows us to select diverse samples with high quality. 
In future work, we intend to extend our method to generative models with high latent dimensions, such as deep hierarchical VAEs \cite{nvae}, or video generative models \cite{tune-a-video} with an additional temporal dimension in latent space. For these models, latent dimension reduction techniques might be a potential solution. 

\textbf{Acknowledgement.} This research was partially supported by NSF grants IIS-2123920 and IIS-2212046.

\clearpage  

\appendix
\section{Appendix-Overview}
In the appendix, we provide additional experiments and analyses. In particular:
\begin{itemize}
\setlength\itemsep{-0.1em}
    \item Section \ref{sec:add_qualitative} provides additional visualizations of the selected samples with the highest/lowest latent density scores across different generative models and datasets.
    \item Section \ref{sec:domain_supp} provides additional results of applying our quality assessment method on other domains and modalities.
    \item Section \ref{sec: trunction} shows the latent density scores for images at different truncation levels.
    \item Section \ref{sec: tsne_visualization} provides visualizations of the latent space of different generative models.
    \item Section \ref{sec: sigma_supp} provides additional analysis on the choice of hyper-parameter $\sigma$ in our score function.
    \item Section \ref{sec: score_justification} justifies the choice of our score function.
    \item Section \ref{sec: efficiency} provides quantitative results of the efficiency gain of our method.
    \item Section \ref{sec: user_study} includes a user study of our method.
\end{itemize}

\begin{figure}[ht!]
\begin{center}
\vspace{-0.2cm}
\includegraphics[clip,width=\columnwidth]{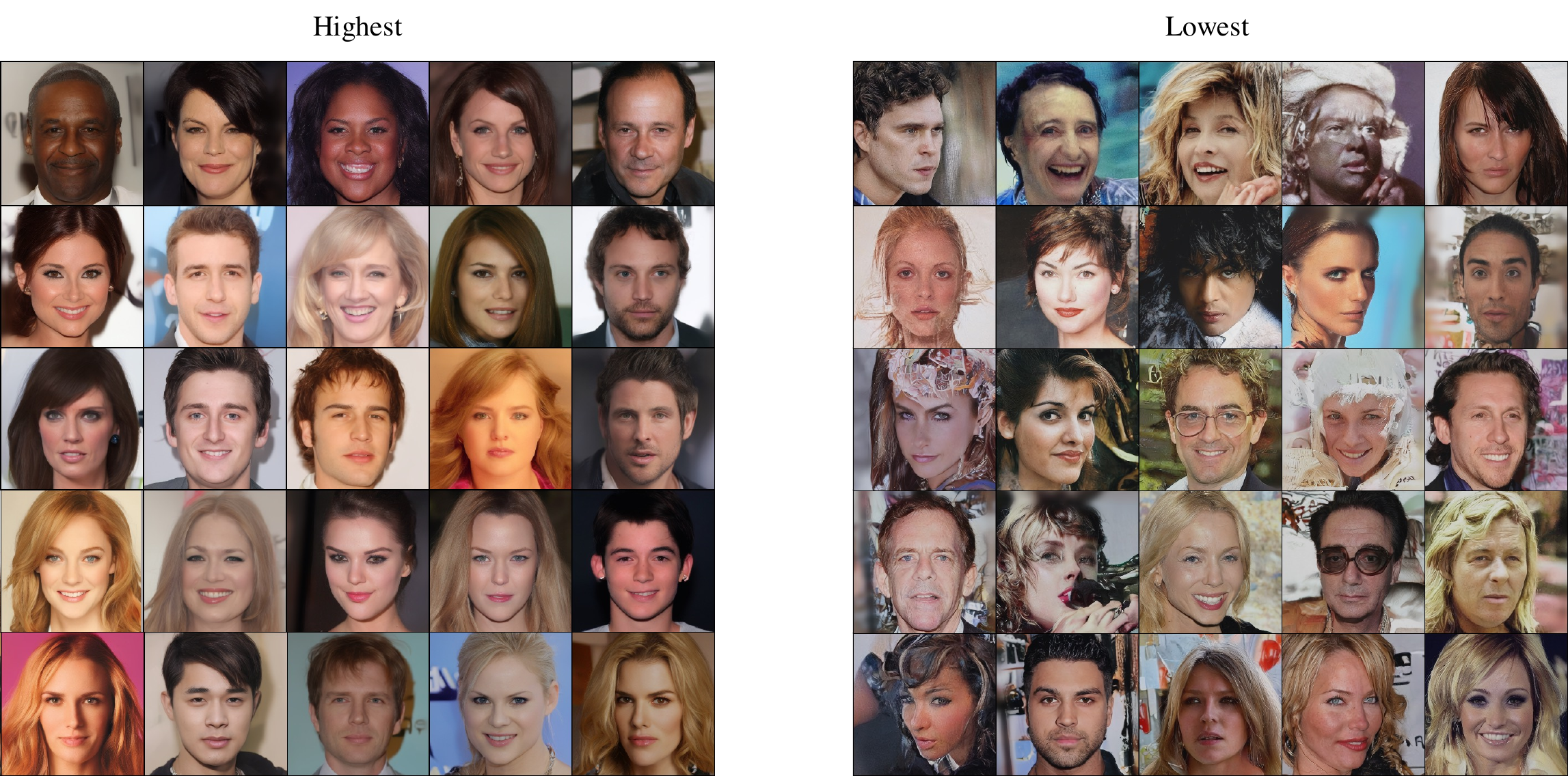}%
\hspace{-0.2cm}
\caption{Samples generated from Latent Diffusion \cite{ldm} trained on Celeba-HQ \cite{celebahq} with the lowest/highest latent density scores.}
\label{fig: celeba_supp}
\end{center}
\end{figure}

\section{Additional Qualitative Results} 
In this section, we show additional selected samples with the highest/lowest latent density scores across different generative models and datasets.
\label{sec:add_qualitative}

\begin{figure}[ht!]
\begin{center}
\vspace{0.1cm}
\includegraphics[clip,width=\columnwidth]{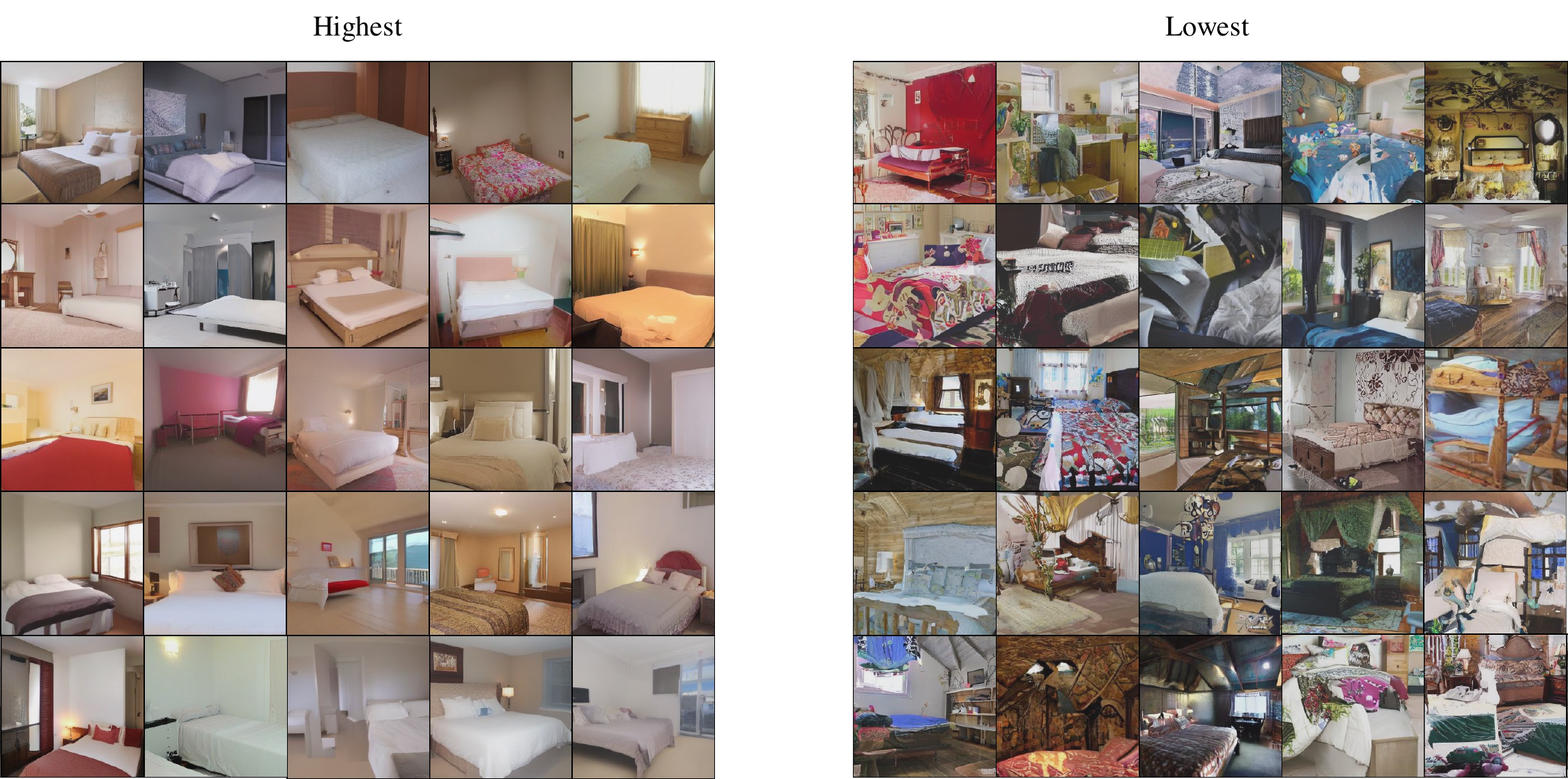}%
\hspace{-0.2cm}
\caption{Samples generated from Latent Diffusion trained on LSUN-Bedrooms \cite{lsun} with the lowest/highest latent density scores.}
\label{fig: bedroom_supp}
\end{center}
\end{figure}

\vspace{0cm}

\begin{figure}[ht!]
\begin{center}
\includegraphics[clip,width=\columnwidth]{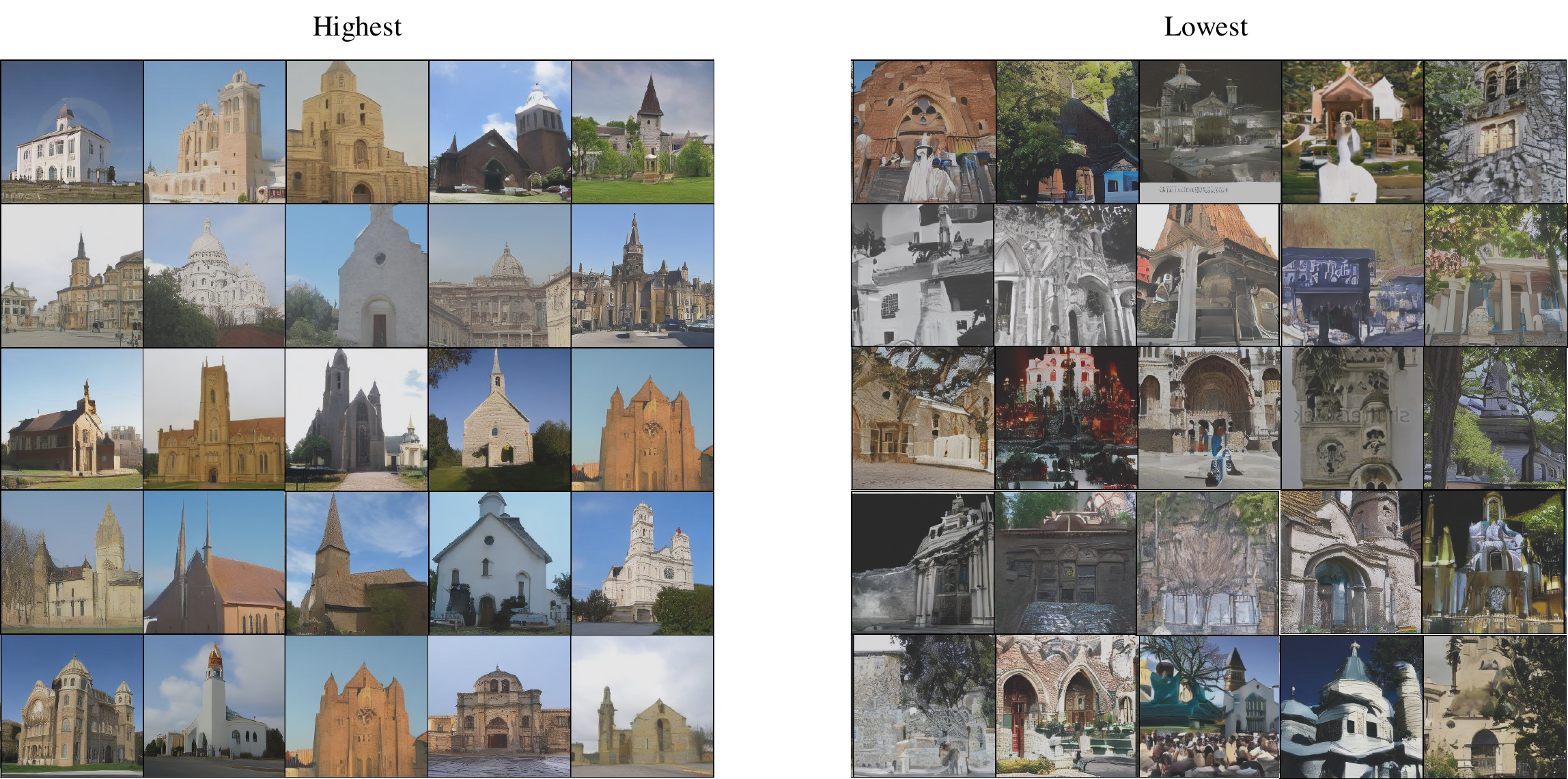}%
\hspace{-0.2cm}
\caption{Samples generated from Latent Diffusion trained on LSUN-Churches \cite{lsun} with the lowest/highest latent density scores.}
\label{fig: churches_supp}
\end{center}
\end{figure}

\begin{figure}[ht!]
\begin{center}
\includegraphics[clip,width=\columnwidth]{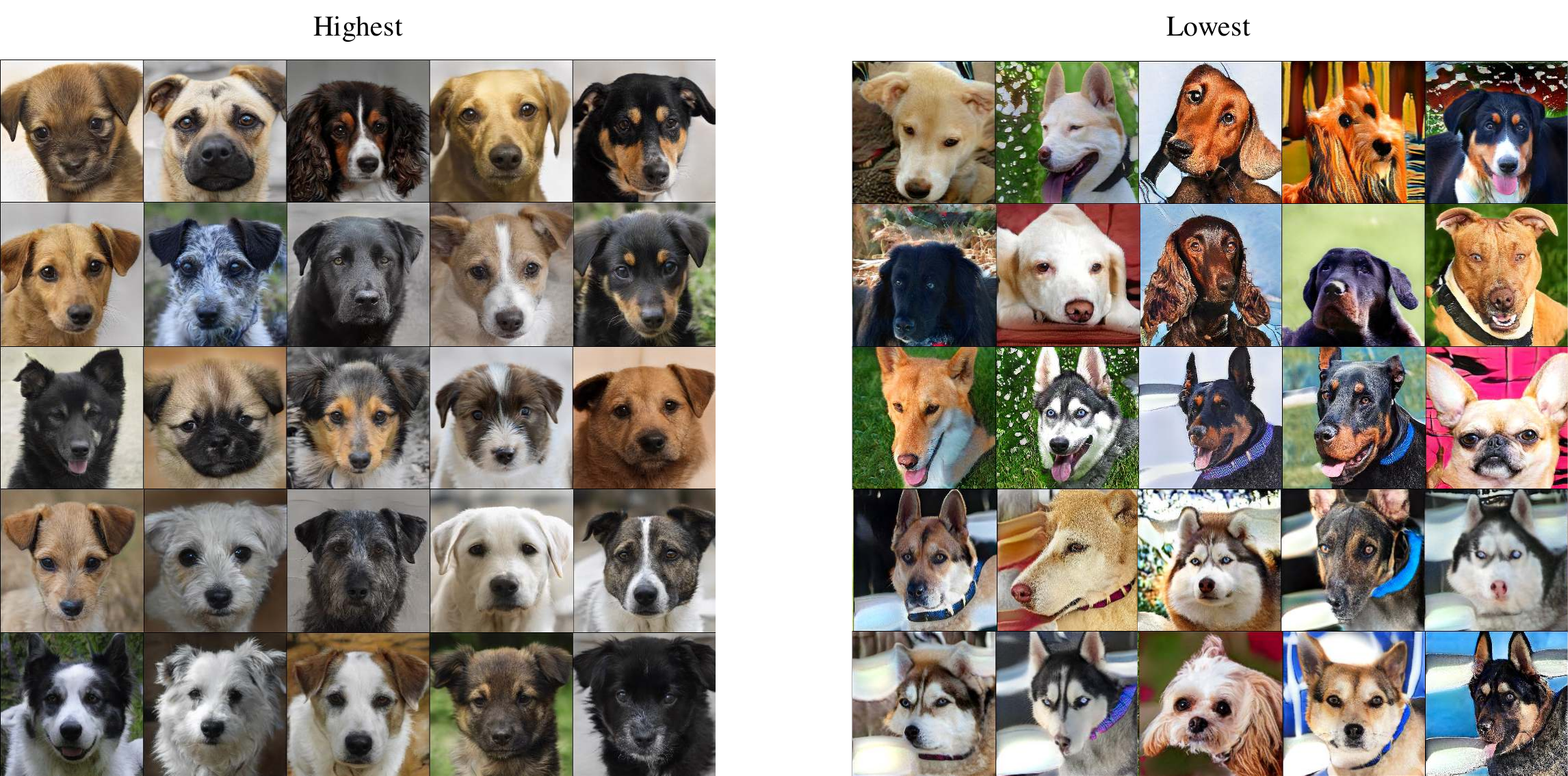}%
\hspace{-0.2cm}
\caption{Samples generated from StyleGAN2-ADA \cite{stylegan2-ada} trained on AFHQ Dog \cite{stargan} with the lowest/highest latent density scores.}
\label{fig: dog_supp}
\end{center}
\end{figure}

\begin{figure}[ht!]
\begin{center}
\includegraphics[width=\columnwidth]{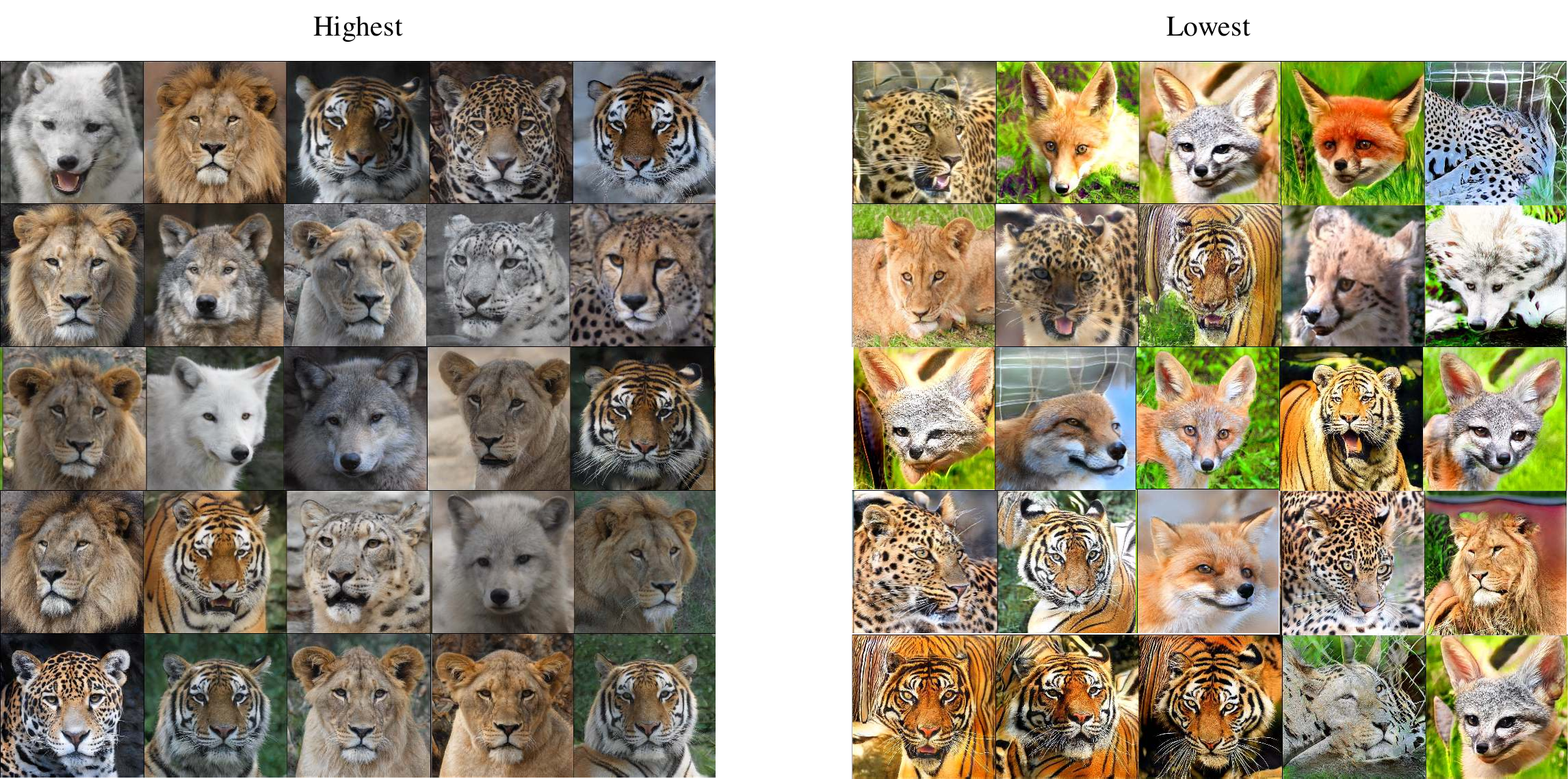}%
\hspace{-0.2cm}
\caption{Samples generated from StyleGAN2-ADA trained on AFHQ Wild \cite{stargan} with the lowest/highest latent density scores.}
\label{fig: wild_supp}
\end{center}
\end{figure}

\section{Additional Results on Other Domains and Modalities}
\label{sec:domain_supp}    
\subsection{Quality Assessment on 3D Shapes}
In Figure \ref{fig: 3d}, we show additional quality assessment results on generated 3D shapes using our proposed latent density score. The 3D shapes are generated via a StyleGAN2-based \cite{stylegan2} 3D shape generation framework, SDF-StyleGAN \cite{zheng2022sdfstylegan} trained on ShapeNet Core V1 \cite{shapenet}. As shown in the figure, the generated samples with high scores have better visual quality with meaningful object structures. The generated samples with low scores, in contrast, exhibit irregular objects shapes and severe geometry distortion. 

\begin{figure*}[htp]
\begin{center}
\subfloat[Airplane]{%
\includegraphics[clip,width=\columnwidth]{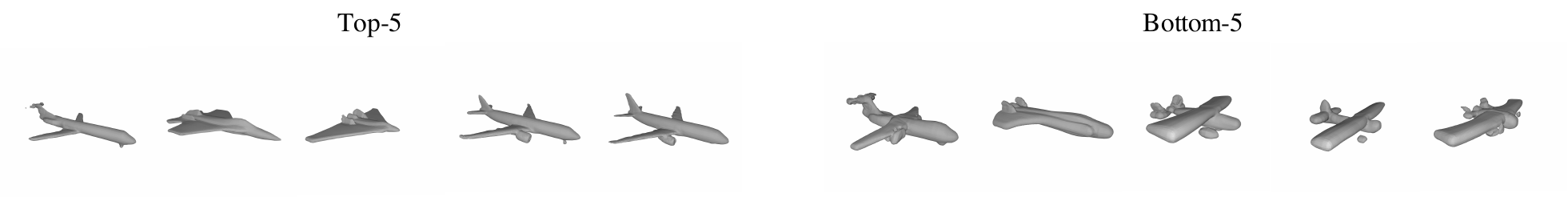}%
}
\vspace{-0.5mm}
\subfloat[Chair]{%
\includegraphics[clip,width=\columnwidth]{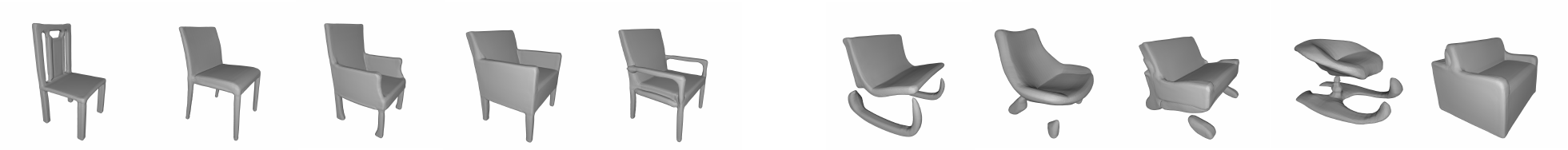}%
}
\vspace{-1mm}
\subfloat[Table]{%
\includegraphics[clip,width=\columnwidth]{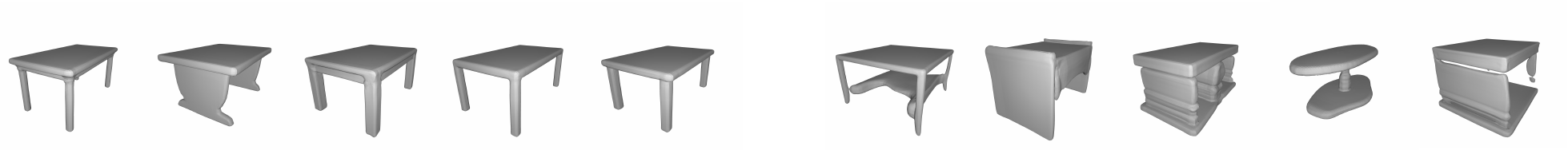}%
}
\vspace{-1mm}
\subfloat[Rifle]{%
\includegraphics[clip,width=\columnwidth]{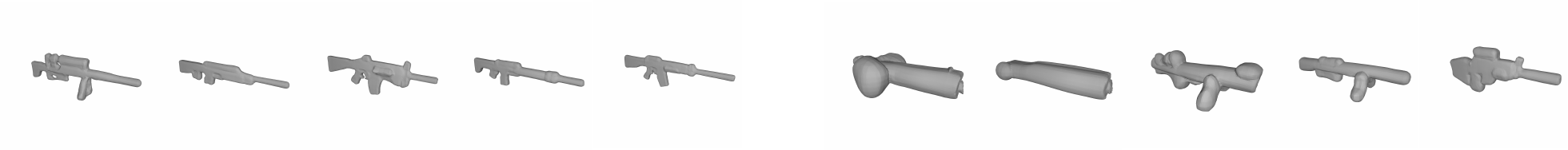}%
}
\caption{Top 5 and Bottom 5 generated 3D shapes for four categories (i.e., airplane, chair, table and rifle) in terms of the proposed latent density score on ShapeNet Core V1 for SDF-StyleGAN.}
\label{fig: 3d}
\end{center}
\end{figure*}

\subsection{Quality Assessment on Non-ImageNet-like Domain Images}

In Figure \ref{fig: brecahad_supp} and Figure \ref{fig: anime_supp}, we show additional selected samples with the highest and lowest latent density scores / realism scores on the BreCaHAD \cite{Aksac2019brecahad} and Danbooru \cite{danbooru} datasets. We see there exhibit clear visual differences between images with the highest and lowest latent density scores on both datasets. However, we do not observe visually distinguishable differences between samples with the highest and lowest realism scores. This is because the pre-trained feature space used by the realism score is not semantically meaningful for non-ImageNet-like domain images. Our proposed latent density score, on the other hand, directly leverages the latent space of generative models and generalizes well across different domains.

\begin{figure}[ht!]
\begin{center}
\hspace{-1cm}\includegraphics[width=\columnwidth]{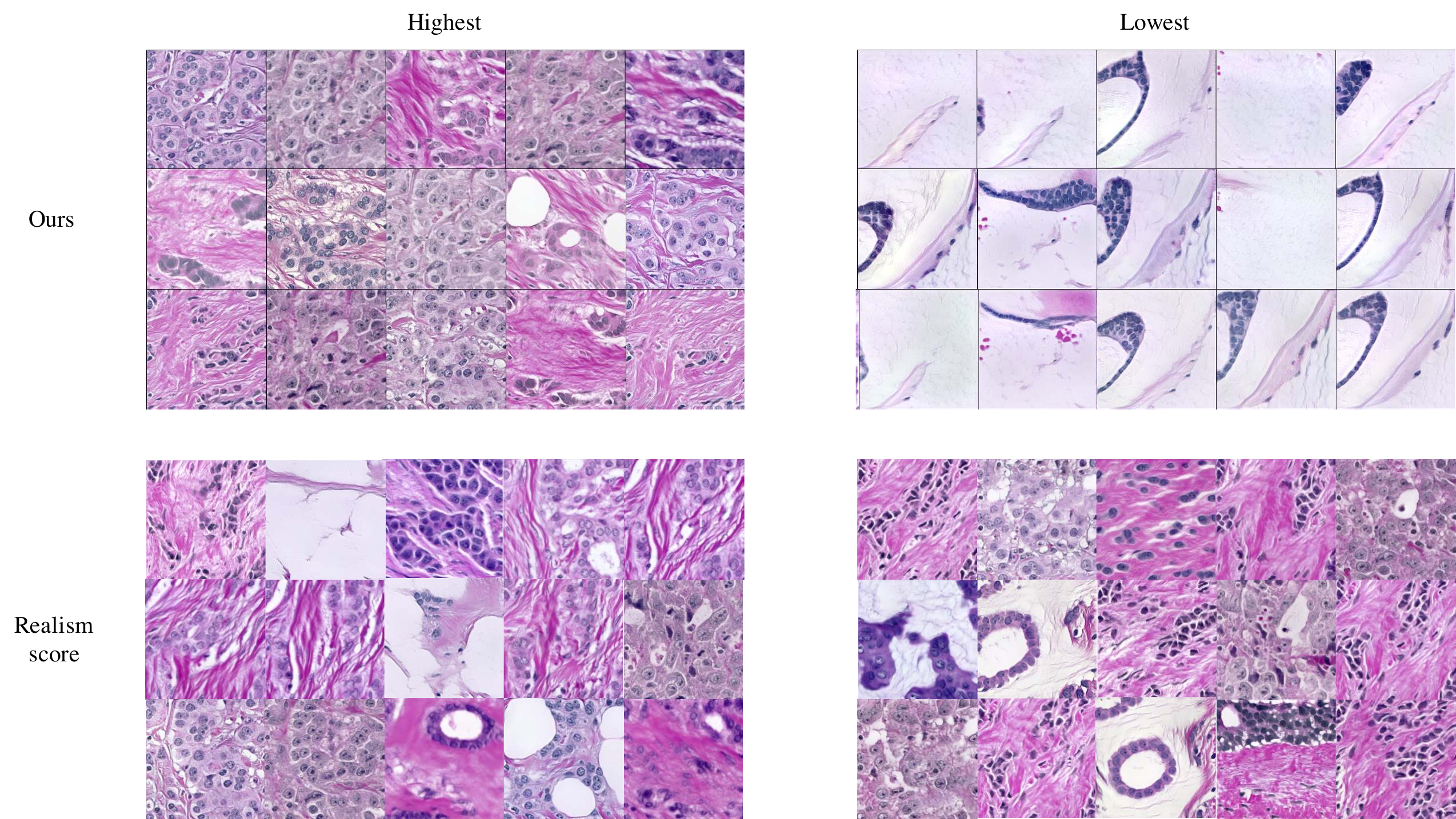}%
\caption{Images with highest and lowest latent density scores / realism scores on BreCaHAD dataset}
\hspace{-0mm}
\label{fig: brecahad_supp}
\end{center}
\end{figure}

\begin{figure}[ht!]
\begin{center}
\vspace{-0.8cm}
\hspace{-1cm}\includegraphics[width=\columnwidth]{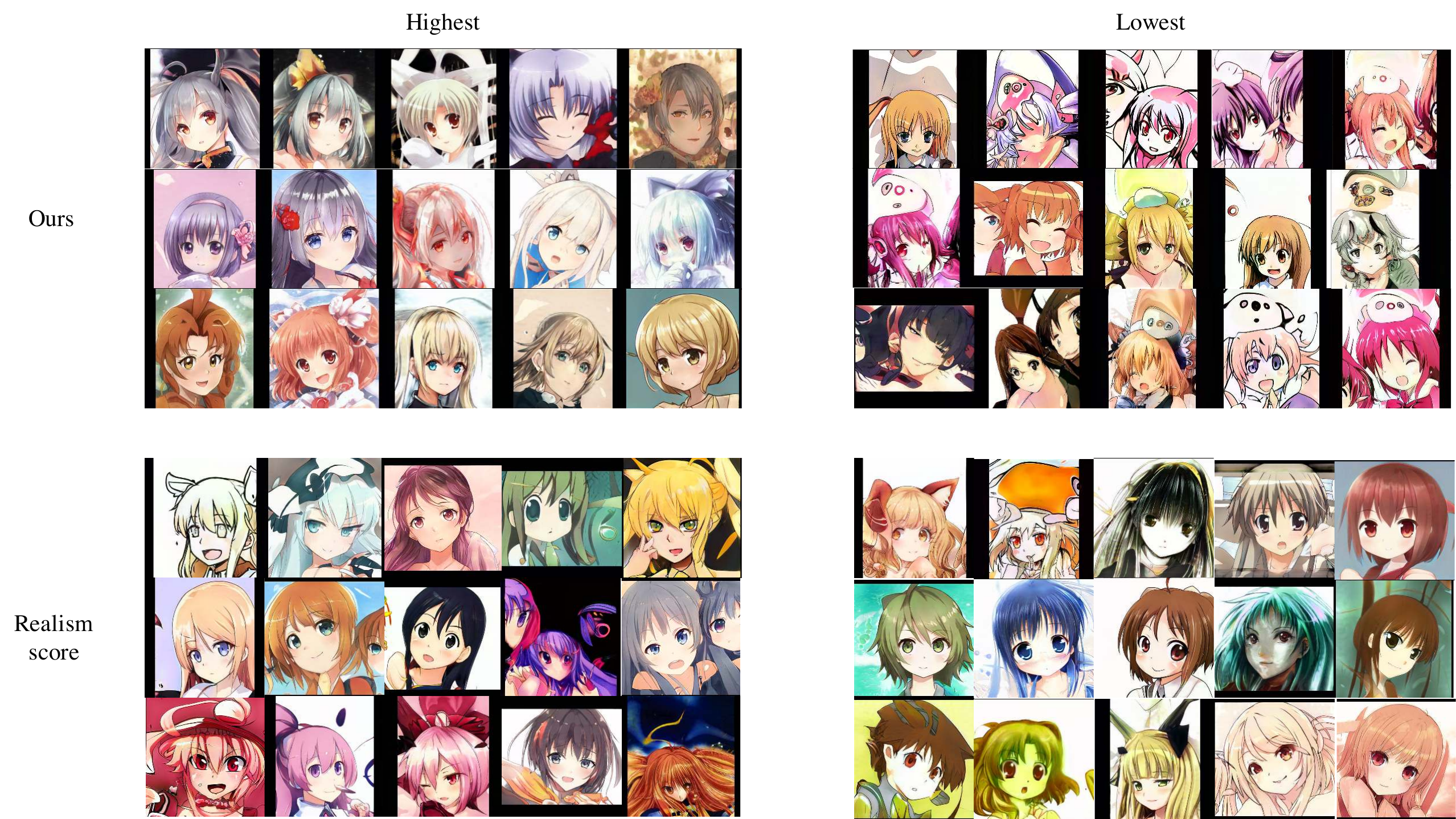}%
\caption{Images with highest and lowest latent density scores / realism scores on Danbooru dataset}
\label{fig: anime_supp}
\end{center}
\end{figure}

\section{Latent Density Scores for Images at Different Truncation Levels}
\label{sec: trunction}

The Truncation trick \cite{stylegan} is used to increase the fidelity of the generated images in GAN-based generative models by moving the latent code towards the mean latent code. The degree of truncation is controlled by the truncation parameter $\psi$, \ie, $\psi = 0$ indicates full truncation using the mean code and $\psi = 1$ indicates no truncation. Figure \ref{fig: truncation} shows the images truncated using different values of $\psi$ and the corresponding latent density scores. As shown in the figure, as $\psi$ goes from $1$ to $0.1$, the perceptual quality of the truncated images gradually increases. For example, the face image in the third row without any truncation shows clear artifacts, while the artifacts are hardly noticeable when $\psi$ is $0.1$. Correspondingly, the latent density score gradually increases as the image quality becomes better. This suggests that the latent density
score is a valid measure for generated sample quality.

\begin{figure}[ht!]
\begin{center}
\hspace{1cm}\includegraphics[width=0.96\columnwidth]{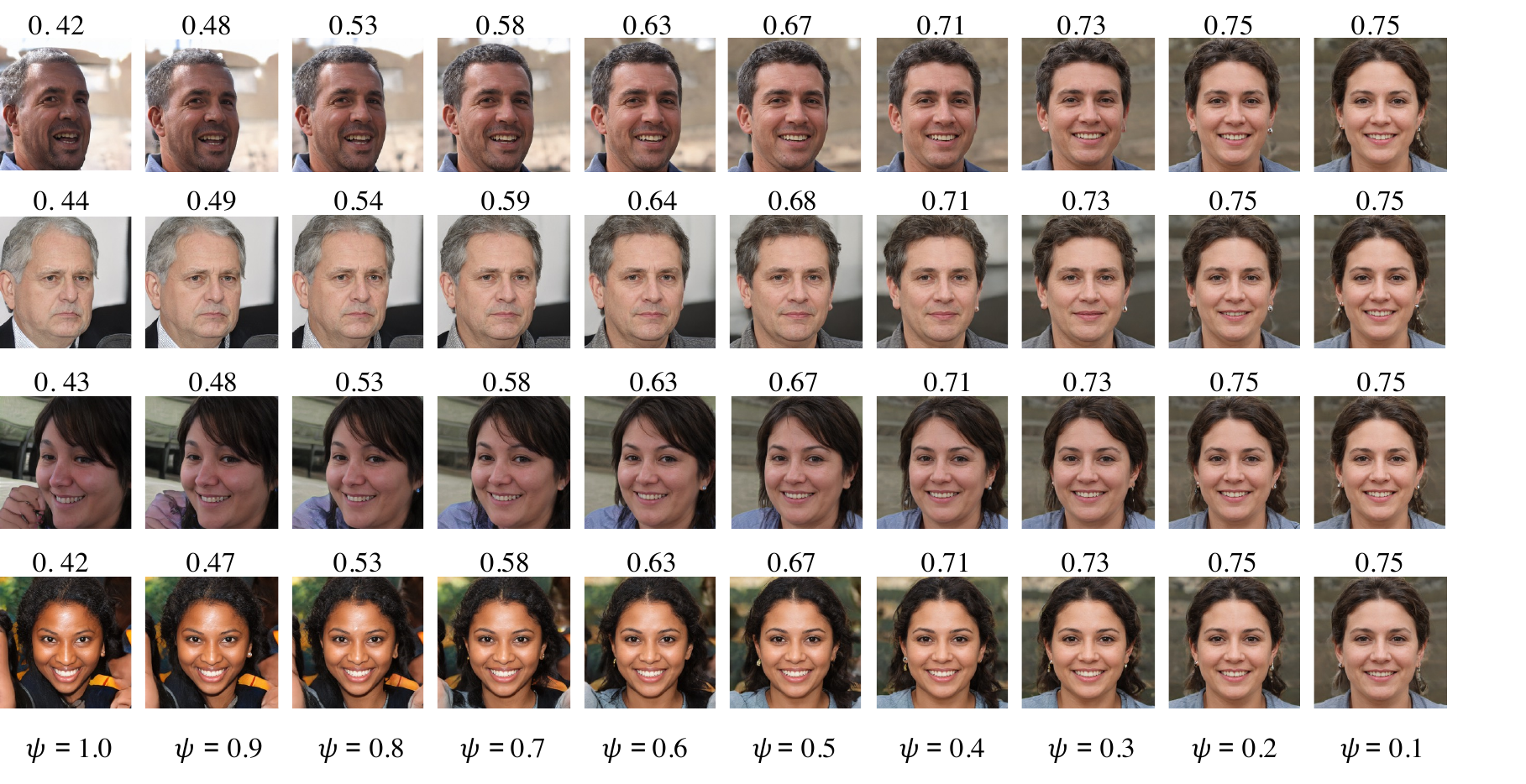}%
\hspace{-0.5cm}
\caption{Latent density scores of images at different truncation levels.}
\label{fig: truncation}
\end{center}
\end{figure}

\begin{figure*}[htp]
\begin{center}
\subfloat[AFHQ Wild]{%
\vspace{0.0cm}
\hspace{0.3cm}\includegraphics[clip,width=\columnwidth]{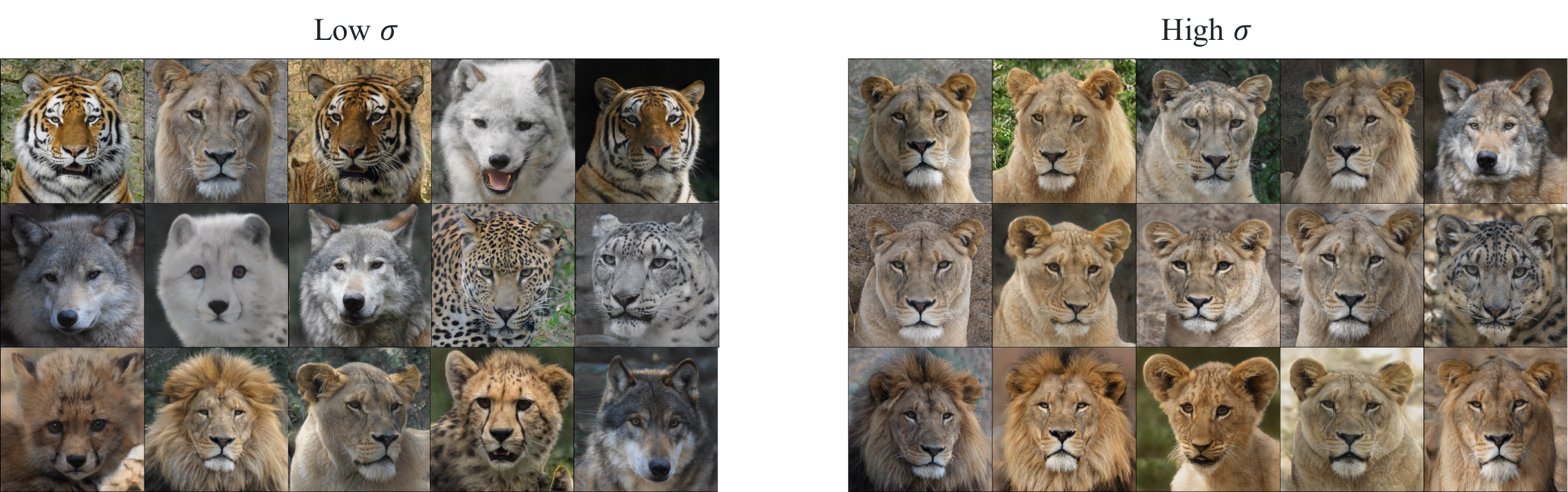}%
}
\vspace{0.0cm}
\subfloat[AFHQ Cat]{%
\vspace{0.0cm}
\hspace{0.3cm}\includegraphics[clip,width=\columnwidth]{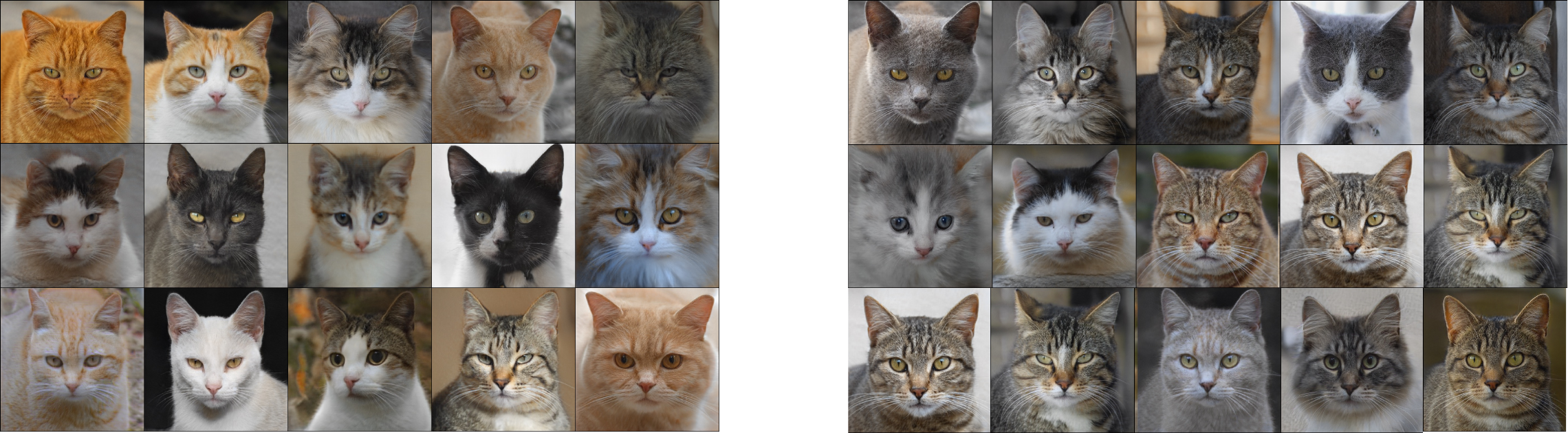}%
}
\caption{Selected high-density samples when using low and high values of $\sigma$ on the AFHQ Wild and AFHQ Cat datasets.}
\vspace{0cm}
\label{fig: sigma}
\end{center}
\end{figure*}

\begin{figure}[ht!]
\begin{center}
\hspace{1.1cm}\includegraphics[width=0.98\columnwidth]{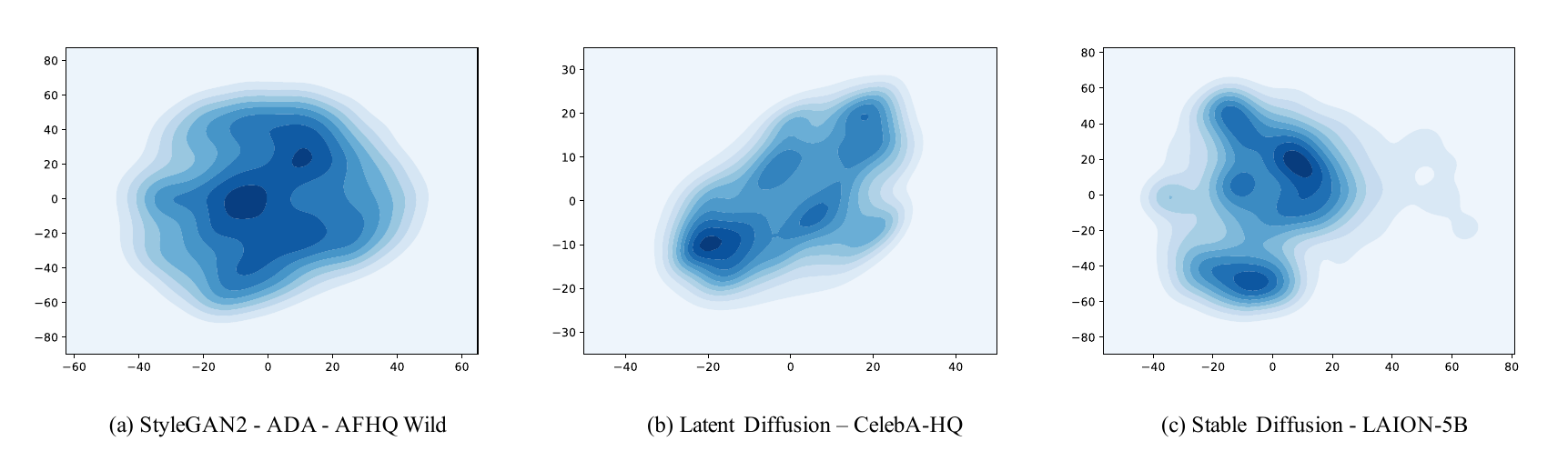}%
\hspace{-0.5cm}
\caption{T-SNE visualization of the latent space of three generative models.}
\label{fig: latent_2d}
\end{center}
\end{figure}

\section{Effect of Hyper-Parameter $\sigma$}
\label{sec: sigma_supp}
As discussed in the main paper, the hyper-parameter $\sigma$ in our score function controls the relative contribution of local density and global density \wrt to the final density score. Applying a small $\sigma$ will increase the chance of selecting the points residing in local clusters as high-density points. Thus, the selected samples are likely to be more diverse.
Figure \ref{fig: sigma} shows additional selected high-score samples when using small and large $\sigma$ values. As shown in the figure, the selected high-density samples when using a small $\sigma$ are more diverse compared to using a large $\sigma$.

\section{T-SNE Visualization of Latent Space}
\label{sec: tsne_visualization}
In Figure \ref{fig: latent_2d}, we show the t-SNE \cite{Maaten2008VisualizingDU} visualization of the latent space of three generative models, \ie, StyleGAN2-ADA \cite{stylegan2-ada} trained on AFHQ Wild \cite{stargan}, Latent Diffusion \cite{ldm} trained on CelebA-HQ \cite{celebahq} and Stable Diffusion \cite{ldm} trained on LAION-5B \cite{Schuhmann2022LAION5BAO}. We randomly select 4k images from the training set of each model and extract their latent embeddings. As shown in the figure, different regions in the latent space show different density patterns: latent codes are densely distributed in some areas, while sparsely distributed in other areas. In fact, we show via extensive experiments that this latent density directly correlates with the quality of generated samples.

\section{Justification of the Score Function}
\label{sec: score_justification}
In our main experiments, we choose Gaussian score function because it is simple, intuitive and effective. In addition, we explore alternative formulations, including two $k$-NN based approaches in \cite{realismscore} and \cite{Han2023rarity}. We show the average realism scores of the top-$k$ samples based on different formulations in Tab. \ref{tab:formulations}.
While our formulation better aligns with the realism score in this case, other functions can also be feasible choices.

\begin{table}[!h] 
\vspace{0pt}
  \centering
\resizebox{0.4\textwidth}{!}{%
  \begin{tabular}{c|ccccc}
    \toprule
    top-$k$ & 200 & 400 & 600 & 800 & 1000  \\
    \midrule
    {\cite{Han2023rarity}} & {1.058} & {1.056} & 1.055 & 1.054 & 1.053 \\ 
    {\cite{realismscore}} & {1.061} & {1.056} & 1.054 & 1.053 & 1.053 \\ 
    Ours & {1.065} & {1.063} & 1.059 & 1.056 & 1.053  \\ 
    \bottomrule
  \end{tabular}} \\ \vspace{0pt}
  \caption{Avg. realism scores of top-$k$ samples. } 
  \vspace{0pt} \label{tab:formulations}
\end{table}

\section{Quantifying Efficiency Improvements}
\label{sec: efficiency}
The efficiency gain of our method mainly comes from bypassing pixel-level image generation. 
For example, generating a single image with an LDM takes $29.378$ secs (with $50$ timesteps on a TITAN RTX GPU). In comparison, our method only requires $0.009$ secs to extract one latent embedding from the LDM's encoder and $0.4$ ms to compute the density score. 

\section{User Study}
\label{sec: user_study}
We conduct a small-scale user study with 12 participants to further verify our method. We ask them to choose the more realistic images between two sets with different scores. Each set contains 9 images. 11 answers (91.7\%) align with the density score. 
%
%
\bibliographystyle{splncs04}
\bibliography{main}
\end{document}